%% file: format_matters_seq2seq.tex
\pdfoutput=1

\documentclass[11pt]{article}

\usepackage{acl}

\usepackage{times}
\usepackage{latexsym}

\usepackage[T1]{fontenc}

\usepackage[utf8]{inputenc}

\usepackage{microtype}

\usepackage{graphicx}
\usepackage{tabularx}
\usepackage{mathtools}
\usepackage{booktabs}
\usepackage{longtable}
\usepackage{tabu}
\usepackage{multirow}
\usepackage{subfigure}
\usepackage{enumitem}
\usepackage{array}

\newcommand{\inputsi}{\emph{input}}
\newcommand{\targetsi}{\emph{target}}

\makeatletter
\newcommand\footnoteref[1]{\protected@xdef\@thefnmark{\ref{#1}}\@footnotemark}
\makeatother

\newcolumntype{H}{>{\setbox0=\hbox\bgroup}c<{\egroup}@{}}

\title{Transforming Sequence Tagging Into A Seq2Seq Task}

\author{
    Karthik Raman\thanks{\,\,\,\,\,Corresponding Author}\,, \,Iftekhar Naim,
    \\
    \textbf{ Jiecao Chen, Kazuma Hashimoto, Kiran Yalasangi, Krishna Srinivasan} \\
    Google Research, Mountain View \\
    \small{ \texttt{\{karthikraman, inaim, chenjiecao, kazumah, kirancy, krishnaps\}@google.com}} \\
}

\begin{document}
\maketitle

\input{0-abstract}

\input{1-introduction}

\input{2-related-work}

\input{3-formats}


\input{4-datasets}


\input{5-exp-setting}

\input{6-results}


\input{7-multilinguality}


\input{X-extractive}


\input{X-robustness}

\input{X-discussions}

\section{Acknowledgements}
We sincerely thank Michael Bendersky, Emily Pitler, Slav Petrov and our anonymous reviewers for their valuable feedback.

\input{limitations}
\input{10-responsible-nlp}

\bibliography{format_matters_seq2seq}
\bibliographystyle{acl_natbib}

\newpage
\clearpage

\appendix
\input{appendix-all}

\end{document}

%% file: 0-abstract.tex
\begin{abstract}
Pretrained, large, generative language models (LMs) have had great success in a wide range of sequence tagging and structured prediction tasks.
Casting a sequence tagging task as a Seq2Seq one requires deciding the formats of the input and output sequences.
However, we lack a principled understanding of the trade-offs associated with these formats (such as the effect on model accuracy, sequence length, multilingual generalization, hallucination).
In this paper, we rigorously study different \emph{formats} one could use for casting input text sentences and their output labels into the \textbf{input} and \textbf{target} (\emph{i.e.,} output) of a Seq2Seq model.
Along the way, we introduce a new format, which we show to to be both simpler and more effective.
Additionally the new format demonstrates significant gains in the multilingual settings -- both zero-shot transfer learning and joint training.
Lastly, we find that the new format is more robust and almost completely devoid of \emph{hallucination} -- an issue we find common in existing formats.
With well over a 1000 experiments studying 14 different formats, over 7 diverse public benchmarks -- including 3 multilingual datasets spanning 7 languages -- we believe our findings provide a strong empirical basis in understanding how we should tackle sequence tagging tasks. 
\end{abstract}

%% file: 1-introduction.tex
\section{Introduction}
\label{sec:introduction}

The advent of powerful generative language models (LMs) such as T5~\cite{raffel2019exploring} and mT5~\cite{xue2020mt5} have unlocked new opportunities.
Recent works have shown great success applying these models to a wide range of sequence tagging and structured prediction tasks \cite{athiwaratkun2020augmented, du2021all, paolini2021structured}.
The key to leveraging these models is to cast the sequence tagging task as a suitable Seq2Seq task -- by transforming the original input text and output labels into the \texttt{input} and \texttt{target} (\emph{i.e.,} output) of a Seq2Seq model.


\begin{table}[h]
\vspace{-0.1in}
\centering
 \resizebox{\textwidth / 2}{!}{
 \begin{tabular}{|c|}
  \hline

\textbf{Slot-filling} example from \textbf{\citet{athiwaratkun2020augmented}}
  \\
  \hline

\textbf{Input:} Add Kent James to the Disney soundtrack

\\

\textbf{Target:} Add [ Kent James | \emph{artist} ] to the [ Disney | \emph{playlist} ] soundtrack

 \\ \hline
 
\textbf{SRL} example from \textbf{ \citet{paolini2021structured}} (predicate = "\underline{sold}")
  \\
  \hline

\textbf{Input:} The luxury auto maker last year [ \underline{sold} ] 1,214 cars in the U.S.

\\

\textbf{Target:}  [ The luxury auto maker |\emph{subject} ] [ last year | \emph{temporal} ] \underline{sold}

\\

[ 1,214 cars | \emph{object} ] 
[ in the U.S. | \emph{location} ]

 \\ \hline

 \end{tabular}
 }

 \caption{Examples of the \textbf{TaggedSpans} approach used in current works to transform NLP problems into the Seq2Seq setting.}
 
 \vspace{-0.1in}
   \label{tab:tagged-spans-examples}
 \end{table}

While the list of NLP problems tackled in this manner are diverse -- Named Entity Recognition, Co-reference resolution, Semantic Role Labeling (SRL), Slot-filling, Entity and Relation extraction to name a few -- the Seq2Seq transformation used across different works is surprisingly similar (with slight variants).
As seen from the examples in Table~\ref{tab:tagged-spans-examples}, existing works use the same Seq2Seq transformation  -- what we refer to as the \textbf{TaggedSpans} \emph{format}.
More specifically, the \emph{target} sentence resembles the input sentence, except for tagged decorators surrounding the corresponding token spans.







While exciting results have been achieved using this \emph{format}, there has been no principled study understanding this format choice.
Towards that end, we make the following contributions in this paper:
\begin{itemize}[noitemsep,topsep=0pt]
    \item Based on rigorous experiments, we derive insights into the performance of the \textbf{TaggedSpans} format (as well as other Seq2Seq transformation formats) along different dimensions.
    
    \item We propose a \textbf{new \emph{Sentinel-based} format} that proves to be not only more effective (achieving significantly higher accuracy across all the benchmarks) but also more efficient than the formats used in existing works.
    
    \item We perform an in-depth study understanding performance of different formats in the \emph{multilingual learning} setting -- with the new \textbf{Sentinel-based} format outperforming other formats by upto \textbf{30-40\%} in some settings.
    
    \item  We demonstrate the \emph{robustness} of different formats to factors such as model size, sequence length and decoding strategy.
    
    \item Lastly, we show that the existing \textbf{TaggedSpans} format (as well as other similar formats) are highly susceptible to hallucinations~\cite{maynez2020faithfulness}, \emph{i.e.,} the proclivity of large LMs adding, modifying, or deleting input words while generating the output sequence. While the \textbf{TaggedSpans} had hallucinated tokens in over \textbf{50\%} of examples in some settings, the proposed \textbf{Sentinel-based} format is virtually \textbf{hallucination-free} (< $0.15\%$ hallucinations).
\end{itemize}

With well over a 1000 experiments involving 14 different formats, 7 datasets (3 multilingual), we believe this to be the most comprehensive study on this topic to-date.

%% file: 2-related-work.tex
\section{Related Work}
Several recent papers have shown the strength of the large Seq2Seq LMs for sequence tagging -- in the context of Named Entity Recognition~\cite{yan-etal-2021-unified-generative, qin2022lfpt, paolini2021structured}, Slot-Labeling for Semantic Parsing~\cite{krishnan2021multilingual, ahmad-etal-2021-intent,du2021all, athiwaratkun2020augmented}, and Semantic Role Labeling~\cite{daza-frank-2018-sequence}. 
\citet{yan-etal-2021-unified-generative} proposed a general Seq2Seq framework for NER tasks, both for disjoint spans and nested overlapping spans.
\citet{ahmad-etal-2021-intent} showed that Seq2Seq models can significantly outperform the encoder-only models for slot-filling tasks. 
\citet{raffel2019exploring} converted a wide range of sequence tagging tasks (e.g., answer span extraction for Question-Answering) into Seq2Seq problems.
All these methods, however, apply fairly similar text-to-text formats in that: they (1) prepend the input with a task prefix and leave rest of it mostly unchanged and (2) use the \textbf{TaggedSpans} format (or slight variants) for the \emph{target} (output) sequence  \emph{i.e.,} copy over input tokens and interlace with the tags for the associated spans.
In this paper, we systematically evaluate this approach, and demonstrate its relative strengths and weaknesses across a large collection of benchmarks.

Recent papers investigated the impact of carefully hand-crafted  input prompts for large language models and showed different prompt templates can significantly change the prediction accuracy, especially in zero-shot or few-shot settings~\cite{reynolds2021prompt, zirui-zeroshot,zhong2021adapting}. For example, \citet{zhong2021adapting} has demonstrated that different prompts describing classification labels can have large impact on the zero-shot classification accuracy. Similar nuances are likely at play for output sequence formats as well, but there hasn't been enough systematic study  to understand the impact of different output format choices. We present empirical evaluations of different input/output formats for sequence tagging tasks, and how they impact sequence tagging accuracy in English-only, multilingual, and zero-shot multilingual settings.

%% file: 3-formats.tex
\begin{table*}[h]
\centering
 \resizebox{\textwidth}{!}{

 \begin{tabular}{|c|llllllll|}
 \hline
 
\textbf{Format} & \multicolumn{8}{l|}{\textbf{Sample \textit{Input} \& \textit{Target}}} \\
  \hline
  \hline

\textbf{Tagged Spans} & 
$\inputsi$: & Add & Kent & James & to & the & Disney & soundtrack \\
 &
$\targetsi$: & <O> Add  </> & <ARTIST> Kent & James </> & <O> to </> & <O> the </> & <PLAYLIST> Disney </> & <O> soundtrack </> \\
\hline

\textbf{Input + Tag} & 
$\inputsi$: & Add & Kent & James & to & the & Disney & soundtrack \\
 &
$\targetsi$: & <O> Add & <ARTIST> Kent & <I-ARTIST> James & <O> to & <O> the & <PLAYLIST> Disney & <O> soundtrack \\
\hline

\textbf{Tag Only} & 
$\inputsi$: & Add & Kent & James & to & the & Disney & soundtrack \\
 &
$\targetsi$: & O & ARTIST & I-ARTIST & O & O & PLAYLIST & O \\
\hline

\textbf{Sentinel + Tag} & 
$\inputsi$: & <extra\_id\_0> Add & <extra\_id\_1> Kent & <extra\_id\_2> James &<extra\_id\_3> to & <extra\_id\_4> the & <extra\_id\_5> Disney & <extra\_id\_6> soundtrack \\
 &
$\targetsi$: & <extra\_id\_0> O & <extra\_id\_1> ARTIST & <extra\_id\_2> I-ARTIST & <extra\_id\_3> O  & <extra\_id\_4> O & <extra\_id\_5> PLAYLIST & <extra\_id\_6> O \\
\hline
\hline
\textbf{Extractive} & 
$\inputsi$: & Add & Kent & James & to & the & Disney & soundtrack \\
\textbf{Tagged Spans} &
$\targetsi$: &  & <ARTIST> Kent & James  &  &  & <PLAYLIST> Disney  &  \\
\hline
\textbf{Extractive} & 
$\inputsi$: & <extra\_id\_0> Add & <extra\_id\_1> Kent & <extra\_id\_2> James &<extra\_id\_3> to & <extra\_id\_4> the & <extra\_id\_5> Disney & <extra\_id\_6> soundtrack \\
\textbf{Sentinel + Tag} &
$\targetsi$: &  & <extra\_id\_1> ARTIST & <extra\_id\_2> I-ARTIST & & & <extra\_id\_5> PLAYLIST &  \\
\hline

\hline
\end{tabular}
 }
\vspace{-0.05in}
 \caption{Example illustrating how the $\inputsi$ and $\targetsi$ differ across the high-level formats studied here.}
 \vspace{-0.1in}
 \label{tab:format-examples}
 \end{table*}

 





\section{The Seq2Seq Transformation}
\label{sec:format}

Applying pretrained generative models on these sequence labeling / structured predictions problems requires creating the $\inputsi$ and $\targetsi$ strings from the original input sentences and labels.
The examples in Table~\ref{tab:tagged-spans-examples} indicate how previous works have tackled this transformation.
In this section, we formalize this transformation and consider different alternatives to the existing \textbf{TaggedSpans} format in how we can create the $\inputsi$ and $\targetsi$ strings.
Note that to simplify the exposition, we assume that the original labels can be represented using the standard \textbf{BIO} (Begin-Inside-Outside) notation -- as is conventional for most of these tasks.

\subsection{The Tagged Spans Format}

This is the approach used by existing works for transforming into a Seq2Seq task. Here the target sentence is an interlaced amalgamation of the input and the associated tag labels. Consider the example shown in Table~\ref{tab:format-examples}.
Here the \emph{"<ARTIST>"} tag precedes the token "Kent" to indicate the start of an "ARTIST" span, and is closed with the \emph{"</>"} marker after the token "James".
In other words, salient token spans are explicitly surrounded by the associated tags \footnote{Note that, to simplify exposition of different formats, we use a slightly different punctuation symbol for the tags than the square brackets used in the examples in Table~\ref{tab:tagged-spans-examples}. This choice does not affect performance though.}

An astute reader may notice that the example from Table~\ref{tab:format-examples} also contains tagged decorators for the "\textbf{O}" (Outside) class as well unlike existing works. This was done to simplify exposition and demonstrate differences from the other formats we study. While we do discuss dropping these Outside tags in Sec.~\ref{sec:format-variants}, it is worth noting that in our empirical analyses we found that having these Outside tag markers actually helps improve performance for the \textbf{TaggedSpans} and other formats (Table~\ref{tab:base-128-256-test-so}).

\textbf{Why this format? }
Given how common this format is in existing works, one may suspect that this choice originates from some meticulous study. However until this paper, there have been no previous efforts exploring other choices, or understanding the relative benefits \& drawbacks of this format.

In particular, one source of concern is the need to \textbf{faithfully} repeat the input as part of the output, \textbf{in addition to} identifying the correct tags for the token spans. This makes the learning task significantly harder (as evidenced by the diminished empirical performance observed in Sec.~\ref{sec:exp-results1}).
Furthermore this copying of text also makes generalization in other languages harder (seen in Sec.~\ref{sec:exp-multilinguality}). 
Additionally the model is harder to interpret since the log-likelihood scores are a combination of the generated input spans and their associated output tags.
Lastly, this copying makes the model susceptible to hallucinations (in over 50\% of examples in some settings  -- Sec.~\ref{sec:hallucinations}).

\subsection{Other Formats}

A slight variant of the \textbf{TaggedSpans} format is the \textbf{Input + Tag} format. As seen from the example in Table~\ref{tab:format-examples}, the output here is an interlacing of the input tokens with the BIO labels \emph{i.e.,} the input token is preceded by its BIO label.
Unfortunately, this format shares the same drawbacks due to the need to accurately repeat the input.

To remedy the issues caused by requiring the input in the target, one solution is to drop the input tokens altogether. This format -- which we call the \textbf{Tag-Only} format -- is the simplest yet as it simply drops the input tokens from the \textbf{Input + Tag} format's target. While this remedies some issues, it makes the learning problem significantly harder as the model now needs to also track token indices implicitly -- a task even harder in non-English languages (as we later demonstrate empirically).

\begin{table*}[h]
\centering
 \resizebox{\textwidth}{!}{
\begin{tabular}{|l|c|c|c|c|c|c|c|c|c|}
 \hline
 
\textbf{Dataset} & \textbf{Lang}
    & \textbf{\# Train} & \textbf{\# Valid} & \textbf{\# Test} 
    & \textbf{Token/ex} & \textbf{Tagged span/ex} & \textbf{\% tokens tagged}
    & \textbf{\# Tag classes} & \textbf{Tag Entropy}\\
  \hline

\multirow{3}{*}{mATIS} & en & 4478 & 500 & 893 &  11.28 & 3.32 & 36.50 & 79 & 3.888 \\

 & hi & 540  & 60 & 893 &  11.39 & 3.14 & 34.21 & 65 & 3.823 \\

 & tr & 540 & 60 & 715 &  10.80 & 3.03 & 33.32 & 63 & 3.847 \\

\hline

SNIPS & en & 13084 & 700 & 700 &  9.00 & 2.60 & 51.33 & 39 & 4.857 \\

\hline

MovieTrivia & en & 7005 & 811 & 1953 &  20.31 & 2.95 & 64.74 & 12 & 2.867 \\

\hline

Movies & en & 8722 & 1019 & 2441 &  10.23 & 2.19 & 38.65 & 12 & 3.145 \\

\hline

Restaurant & en & 6845 & 789 & 1516 &  9.25 & 2.01 & 38.02 & 8 & 2.809 \\

\hline

\multirow{6}{*}{mTOP} & en & 15667 & 2235 & 2235 &  7.61 & 1.71 & 44.18 & 74 & 4.681 \\

 & de & 13424 & 1815 & 3549 &  7.97 & 1.70 & 42.73 & 74 & 4.613 \\

 & es & 10933 & 1527 & 2998 &  9.01 & 1.63 & 42.95 & 70 & 4.522 \\
 
 & fr & 11814 & 1577 & 3193 &  9.03 & 1.61 & 39.08 & 73 & 4.604 \\
 
 & hi & 10933 & 2012 & 2789 &  8.28 & 1.63 & 38.71 & 72 & 4.714 \\
 
 & th & 10759 & 1671 & 2765 &  9.21 & 1.58 & 48.37 & 73 & 4.608 \\

\hline

\multirow{3}{*}{mTOD} & en & 30506 & 4178 & 8615 &  7.26 & 1.69 & 43.46 & 15 & 2.434 \\

 & es & 3616  & 1981 & 3043 &  7.46 & 1.55 & 52.43 & 11 & 2.323 \\

 & th & 2154  & 1235 & 1692 &  7.46 & 1.55 & 52.43 & 11 & 2.323 \\

\hline

\end{tabular}
 }
 \vspace{-0.1in}
 \caption{Statistics for the different datasets, including split sizes as well as per-examples averages of tokens/example, tagged spans/example. A higher \emph{\% tokens tagged} value indicates more tokens being part of tagged spans. The number of tag classes and tag entropy indicate the difficulty of identifying a tag for any given span. Note that the last 5 columns are computing using only the training sets.}
 \vspace{-0.15in}
 \label{tab:dataset-stats}
 \end{table*}

\subsection{The proposed Sentinel+Tag Format}

To avoid these issues with the aforementioned formats, we would like a format that:
\begin{itemize}[noitemsep,topsep=0pt]
    \item Avoids having to repeat the input to simplify decoding and avoid hallucination.
    \item Allows for easier way of tracking tokens (and indices) and associating them with their tags.
    \item Does so in a language-agnostic manner so as to enable multilingual generalization.
\end{itemize}
To achieve this, the \textbf{key insight} we had was \emph{to slightly modify the input to simplify tracking tokens}.
In particular, we proposed using text-independent, language-independent \textbf{Sentinel tokens} -- see example in Table~\ref{tab:format-examples}.
Introducing these sentinels in the input, allows us to avoid repeating the original text tokens in the target, and instead use the sentinel tokens\footnote{In
this work we use one sentinel per token. In particular, the sentinel we used for token number $k$ is \texttt{<extra\_id\_k>} -- since these are present in the mT5 vocabulary as single tokens.}
associated with each input token to help learn the corresponding BIO tag.
By virtue of the resulting targets being independent of the text tokens and language-agnostic, the \textbf{Sentinel+Tag} format can generalize well and be largely immune to hallucination.
Furthermore, unlike the \textbf{Tag-Only} approach, the learning problem is simplified as the sentinels help uniquely ground tokens (spans) and the associate tags.
Lastly, while the input is slightly longer, this is completely offset by the massive reduction in output sequence lengths -- see Table~\ref{tab:seq-len-stats}.

While questions may arise about the effect of imperfect tokenization (e.g. in non-space separated languages like Thai), our empirical study in Section~\ref{sec:exp-multilinguality} spanning 7 languages aims to answer this.

\subsection{Variants and Simplifications}
\label{sec:format-variants}

We also investigated further variants of these formats, that simplify the target texts.
One such simplification pertains to the \emph{Inside} tags. Rather than having to produce the complete tag (e.g. "\emph{I-ARTIST}"), we also evaluated variants where the target simply uses the tag "\emph{I}" regardless of the opened tag. When this simplification is used, we refer to it as \textbf{Simplified Inside (SI)}.

Another such simplification pertains to the \emph{Outside} tags. Rather than emitting these tags, we also evaluated variants where the Outside tag was omitted from the target. 
Examples for these variants can be found in Table~\ref{tab:format-variant-examples} in the Appendix.
We refer to this simplification as \textbf{Simplified Outside (SO)}.

Note that the rest of the paper assumes non-nested spans for simplicity. However most of these above mentioned formats can easily support nested or overlapping spans (Example: By having multiple tags for each opened span after the corresponding token / sentinel in the \textbf{Tag + Input} or \textbf{Sentinel + Tag} approaches).

\subsection{Formats for Extractive Applications}
\label{sec:format-extractive}

The approaches discussed so far allowed us to map the predicted label to the exact span occurrence in the input string.
However in some scenarios this input-alignment is not necessary. In particular, for certain tasks and applications we primarily care about extracting only the labeled tags and associated phrases.
In such cases we can further simplify the target as seen in the target of the \textbf{Extractive Tagged Spans} (abbrv. \textbf{ETS}) in Table~\ref{tab:format-examples}.
This format, which is used in the T5 paper \citep{raffel2019exploring}, simply outputs only the labeled spans.
While this simplifies the task, it comes at the cost of not being able to map spans precisely\footnote{While some heuristic matching can be used, it is hard to do so in a multilingual-friendly manner. Additionally multiple mentions with different semantic meanings can be problematic (e.g. “who plays harry potter in the harry potter movies”)} back to the input for non-extractive tasks.

We can analogously define \textbf{Extractive Sentinel+Tag} (abbrv. \textbf{ES+T}) formats for this extractive setting (see last row of  Table~\ref{tab:format-examples}). Note that, unlike the existing ETS approaches, the sentinel approach allows us to precisely map labels back to the original input token indices and thus can be used seamlessly for \textbf{both} extractive and non-extractive tasks. However, to simplify exposition of the paper, we separate the ES+T and ES+T \textbf{Simplified} (abbrv. \textbf{ES+T(S))} -- which drops Inside tags (see example in Table~\ref{tab:format-variant-examples}) -- approaches and only discuss these with the other extractive formats.

%% file: 4-datasets.tex
\section{Datasets}
\label{sec:datasets}

To gain a comprehensive understanding of how these different formats perform, we ran experiments on 7 different public datasets -- 4 monolingual and 3 multilingual spanning 7 languages from different language groups.
The monolingual datasets include: SNIPS~\cite{coucke2018snips}, MIT Movie Trivia\footnote{\label{footnote:mit-data}All the MIT datasets were downloaded from https://groups.csail.mit.edu/sls/downloads/.}, MIT Movie corpus, and MIT Restaurant corpus.
The multilingual datasets include: mATIS \cite{upadhyay2018almost} -- a multilingual generalization of the ATIS dataset\cite{price1990evaluation}, mTOP \cite{li2021mtop},  and mTOD \cite{schuster2018cross}.
These datasets span different domains -- such as cinema, dining, travel, dialog $\dots$ -- and contain different kinds of texts -- including queries, reviews, and conversations.
For instance, mATIS, SNIPS, mTOP, and mTOD contain conversational queries or dialog-like texts, and involve slot labelling for semantic parsing.
The MIT Movie Trivia, MIT Movies, and MIT Restaurants datasets, on the other hand, are widely used for Named Entity Recognition (NER).

As seen from the statistics listed in Table~\ref{tab:dataset-stats}, these datasets tend to have different properties.
MIT Movie Trivia dataset tends to have longer sentences, longer tagged spans, and more named entities but fewer number of output tag categories.
On the other hand, mATIS and mTOP have relatively shorter inputs but contain over 70 different output tags.
The mTOD dataset has shorter inputs and fewer output categories and is the largest training set (> 30K for English), making it a relatively easier task.
Finally, the high entropy of tags (as in the mTOP and SNIPS) indicates a higher degree of difficulty with fairly ambiguous categories.
Together these datasets provide a broad and diverse collection of public benchmarks for evaluating different Seq2Seq formats.


%% file: 5-exp-setting.tex
\section{Empirical Setting}
\label{sec:exp-setting}

We studied different input/output formats using the mT5 pretrained  models \cite{xue2020mt5}, which have been used to obtain state-of-the-art results on multiple public benchmarks.
In particular, we used the \textbf{Base}-sized model for the majority of our experiments as it enabled us to maximize our experimentation given our compute budget, while still providing for a very powerful initialization.
However, results and trends largely stay the same for other model sizes as we discussed in Sec~\ref{sec:exp-robustness}.

In all cases, we trained these models using the T5X\footnote{https://github.com/google-research/t5x} code for 5000 steps -- selecting the checkpoint with the highest validation set sequence accuracy.
As recommended in the T5 paper we used the default learning rate of 0.001 for all our experiments, and used a batch size of 128.
We appended EOS tokens at the ends of all input and target formats for consistency.
Lastly, we used an input sequence length of 128 and target sequence length of 256, while also using packing.
The impact of training sequence length on accuracy and speed is discussed further in Sec~\ref{sec:exp-robustness}.

To evaluate the different Seq2Seq formats, we primarily relied on two metrics:
\begin{enumerate}[noitemsep,topsep=0pt]
    \item \textbf{Perfect}: This reflects the \% of times, an example is parsed perfectly \emph{i.e.,} all tagged spans are identified correctly with the right tag.
    \item \textbf{F1}: Using CoNLL style evaluation of tagged spans, we report Micro F1 (Macro-F1 trends were consistent with Micro-F1 findings).
\end{enumerate}
One important aspect to note is that the outputs of all the formats are different.
Hence they need to be evaluated differently.
In particular, formats that repeat the input (\textbf{Tagged Spans}, \textbf{Input + Tag}) are more prone to hallucination.
To try and decouple this effect where possible, we evaluated the  \textbf{Tagged Spans} and \textbf{Input + Tag} format by checking the token indices of the tagged spans -- rather than the generated token strings. While this leads to slightly higher metrics, it allows for a more fair and hallucination-free evaluation against approaches like \textbf{Tag-Only} and \textbf{Sentinel + Tag}.


%% file: 6-results.tex

\begin{table*}[h]
\centering
 \resizebox{\textwidth}{!}{
 \begin{tabular}{|l|cc||cc|cc|cc|cc||cc|cc|cc|}
  \hline

\multirow{2}{*}{\textbf{Dataset}} & \multicolumn{2}{c||}{\textbf{mBERT} (and SI)}  & \multicolumn{2}{c|}{\textbf{Tag Only}} & \multicolumn{2}{c|}{\textbf{Input + Tag}} & \multicolumn{2}{c|}{\textbf{Tagged Spans}} & \multicolumn{2}{c||}{\textbf{Sentinel + Tag}} & \multicolumn{2}{c|}{\textbf{Tag Only (SI)}} & \multicolumn{2}{c|}{\textbf{Input + Tag (SI)}} & \multicolumn{2}{c|}{\textbf{Sentinel + Tag (SI)}}
 \\
& \textbf{Perfect} & \textbf{F1} & \textbf{Perfect} & \textbf{F1} & \textbf{Perfect} & \textbf{F1} & \textbf{Perfect} & \textbf{F1} & \textbf{Perfect} & \textbf{F1} & \textbf{Perfect} & \textbf{F1} & \textbf{Perfect} & \textbf{F1} & \textbf{Perfect} & \textbf{F1}
 \\
 \hline

\textbf{mATIS}(en) & 85.11 & 93.65 & 88.17 & 93.83 & 88.73 & 95.40 & 88.95 & 95.55 & \textbf{89.81}\textsuperscript{\textdagger} & \textbf{95.91}\textsuperscript{\textdagger} & 88.65 & 94.77 & 88.76 & 95.32 & \textbf{90.07}\textsuperscript{\textdagger} & \textbf{95.96}\textsuperscript{\textdagger}
\\ 
 & (84.88) & (93.25) & $\pm$ 0.37 & $\pm$ 0.28 & $\pm$ 0.46 & $\pm$ 0.21 & $\pm$ 0.43 & $\pm$ 0.16 & $\pm$ 0.09 & $\pm$ 0.05 & $\pm$ 0.78 & $\pm$ 0.37 & $\pm$ 0.56 & $\pm$ 0.34 & $\pm$ 0.47 & $\pm$ 0.13
 \\ \hline
\textbf{SNIPS} & 86.57 & 93.66 & 85.38 & 92.24 & 87.05 & 94.30 & 87.14 & 94.28 & \textbf{90.14}\textsuperscript{\textdagger} & \textbf{95.47}\textsuperscript{\textdagger} & 89.10 & 94.73 & 87.76 & 94.47 & \textbf{89.81}\textsuperscript{\textdagger} & \textbf{95.53}\textsuperscript{\textdagger}
\\ 
  & (88.29) & (94.34) & $\pm$ 0.54 & $\pm$ 0.42 & $\pm$ 0.36 & $\pm$ 0.20 & $\pm$ 0.31 & $\pm$ 0.09 & $\pm$ 0.70 & $\pm$ 0.40 & $\pm$ 0.36 & $\pm$ 0.28 & $\pm$ 0.78 & $\pm$ 0.29 & $\pm$ 0.07 & $\pm$ 0.04
 \\ \hline
\textbf{MovieTrivia} & 29.80 & 64.69 & 22.65 & 59.10 & 39.84 & 72.58 & \textbf{40.21} & 72.19 & 39.43 & \textbf{72.67} & 34.27 & 68.41 & 39.73 & 72.76 & \textbf{39.85} & \textbf{73.01}
\\ 
  & (29.60) & (64.40) & $\pm$ 3.20 & $\pm$ 2.48 & $\pm$ 0.47 & $\pm$ 0.20 & $\pm$ 0.09 & $\pm$ 0.22 & $\pm$ 0.68 & $\pm$ 0.58 & $\pm$ 0.23 & $\pm$ 0.25 & $\pm$ 0.30 & $\pm$ 0.14 & $\pm$ 0.58 & $\pm$ 0.27
 \\ \hline
\textbf{Movies} & 60.59 & 80.87 & 71.24 & 86.79 & 71.94 & 87.26 & 71.42 & 86.95 & \textbf{73.17} & \textbf{87.79} & 72.06 & 87.18 & 72.12 & 87.47 & \textbf{72.74} & \textbf{87.56}
\\ 
  & (61.41) & (81.40) & $\pm$ 0.18 & $\pm$ 0.06 & $\pm$ 0.27 & $\pm$ 0.25 & $\pm$ 0.36 & $\pm$ 0.13 & $\pm$ 0.54 & $\pm$ 0.34 & $\pm$ 0.58 & $\pm$ 0.30 & $\pm$ 0.02 & $\pm$ 0.13 & $\pm$ 0.95 & $\pm$ 0.48
 \\ \hline
\textbf{Restaurant} & 50.06 & 70.25 & 61.28 & 78.63 & 62.12 & \textbf{80.14} & 61.68 & 79.87 & \textbf{62.53} & 80.02 & 61.87 & 79.38 & 62.42 & \textbf{80.42} & \textbf{62.93} & 80.39
\\ 
  & (50.79) & (71.03) & $\pm$ 0.19 & $\pm$ 0.26 & $\pm$ 0.14 & $\pm$ 0.19 & $\pm$ 0.69 & $\pm$ 0.35 & $\pm$ 0.65 & $\pm$ 0.46 & $\pm$ 0.19 & $\pm$ 0.07 & $\pm$ 0.22 & $\pm$ 0.15 & $\pm$ 0.30 & $\pm$ 0.31
 \\ \hline
\textbf{mTOP}(en) & 81.16 & 88.48 & 81.98 & 88.50 & 84.34 & 90.80 & 84.22 & 90.82 & \textbf{85.58}\textsuperscript{\textdagger} & \textbf{91.68}\textsuperscript{\textdagger} & 83.06 & 89.59 & 84.88 & 91.20 & \textbf{86.56}\textsuperscript{\textdagger} & \textbf{92.28}\textsuperscript{\textdagger}
\\ 
  & (81.21) & (88.69) & $\pm$ 0.34 & $\pm$ 0.34 & $\pm$ 0.45 & $\pm$ 0.39 & $\pm$ 0.66 & $\pm$ 0.46 & $\pm$ 0.58 & $\pm$ 0.42 & $\pm$ 0.48 & $\pm$ 0.31 & $\pm$ 0.21 & $\pm$ 0.06 & $\pm$ 0.69 & $\pm$ 0.44
 \\ \hline
\textbf{mTOD}(en) & 92.05 & 95.71 & 92.53 & 95.97 & 92.39 & 96.01 & 92.13 & 95.90 & \textbf{92.69} & \textbf{96.21} & 93.09 & 96.34 & 92.66 & 96.17 & \textbf{93.19} & \textbf{96.42}
\\ 
  & (93.32) & (97.12) & $\pm$ 0.10 & $\pm$ 0.05 & $\pm$ 0.07 & $\pm$ 0.04 & $\pm$ 0.08 & $\pm$ 0.08 & $\pm$ 0.13 & $\pm$ 0.08 & $\pm$ 0.06 & $\pm$ 0.04 & $\pm$ 0.03 & $\pm$ 0.01 & $\pm$ 0.17 & $\pm$ 0.08
 \\ \hline
\hline
\textbf{Average} & 69.33 & 83.90 & 71.89 & 85.01 & 75.20 & 88.07 & 75.11 & 87.94 & \textbf{76.19}\textsuperscript{\textdagger} & \textbf{88.54}\textsuperscript{\textdagger} & 74.59 & 87.20 & 75.48 & 88.26 & \textbf{76.45}\textsuperscript{\textdagger} & \textbf{88.73}\textsuperscript{\textdagger}
\\ 
  & (70.05) & (84.32) & $\pm$ 0.25 & $\pm$ 0.22 & $\pm$ 0.18 & $\pm$ 0.11 & $\pm$ 0.11 & $\pm$ 0.03 & $\pm$ 0.19 & $\pm$ 0.16 & $\pm$ 0.15 & $\pm$ 0.11 & $\pm$ 0.19 & $\pm$ 0.08 & $\pm$ 0.26 & $\pm$ 0.16
 \\ \hline

 \hline
 \end{tabular}
 }
\vspace{-0.1in}
 \caption{Results for how the different Seq2Seq formats perform on the English benchmarks. Metrics are averaged over 3 runs and reported (with  standard deviation). mBERT results include both w/o and with Simplified Inside. The \textsuperscript{\textdagger} symbol indicates 99+\% significant improvement (per the z-test) against \textbf{all} non-sentinel approaches.}
\vspace{-0.15in}
   \label{tab:base-128-256-test}
 \end{table*}


\begin{table}[!h]
\centering
 \resizebox{\textwidth / 2}{!}{
 \begin{tabular}{|l|c|c|c|}
  \hline

\textbf{Dataset} & \textbf{Input + Tag} & \textbf{Tagged Spans} & \textbf{Sentinel +} 
\\
 & \textbf{(SI, SO)} & \textbf{(SO)} & \textbf{Tag (SI, SO)} 
  \\
  \hline
\multicolumn{4}{|c|}{\textbf{English-only benchmarks}} 
\\ \hline
\textbf{Average} & 74.35 & 74.85 & 75.68 
\\
($\Delta$ vs. non-SO) & -1.13 & -0.26 & -0.77 
 \\ \hline
\multicolumn{4}{|c|}{\textbf{Multilingual Zero-Shot}}
\\ \hline
\textbf{Average} (non-en) & 21.33 & 21.05 & 44.76 
\\
($\Delta$ vs. non-SO) & -15.14 & -10.24 & -2.35 
 \\ \hline
\multicolumn{4}{|c|}{\textbf{Multilingual Joint}}
\\ \hline
\textbf{Average} (non-en) & 73.27 & 72.93 & \textbf{79.01} 
\\
($\Delta$ vs. non-SO) & -1.62 & 0.33 & 0.23 
 \\ \hline
 \end{tabular}
 }

 \caption{\textbf{Perfect} metric scores for the variant Seq2Seq formats that modify or drop the Outside tag (along with the  performance difference due to this simplification).}
   \label{tab:base-128-256-test-so}
  \vspace{-0.2in}
 \end{table}

\section{Performance on English-benchmarks}
\label{sec:exp-results1}

The first question we looked to answer was: \textit{How well do these Seq2Seq approaches perform compared with the previous encoder-only approaches?}.
To do so, as a baseline we used the previous benchmark setting mBERT \cite{devlin2018bert} model -- which is an encoder-only model having a similar size as the T5 base model's encoder.
We evaluated the models on the simpler \textit{English-only} setting to start with.
The results on all 7 English benchmarks can be found in Table~\ref{tab:base-128-256-test}.
Perhaps unsurprisingly, we find that the Seq2Seq approaches significantly outperform the encoder-only mBERT approach on all datasets. This result is consistent with prior papers~\cite{ahmad-etal-2021-intent, raffel2019exploring}.
All but the simplest Tag-Only Seq2Seq format outperform the mBERT baseline on all the datasets.
This conclusively demonstrates the value of modeling these problems using the encoder-decoder models.

The next question we looked to answer was: \textit{How do the different basic formats described in Sec~\ref{sec:format} compare against each other?} In particular, we wanted to understand whether not repeating the input tokens hurts performance.
As seen from the third to sixth columns in Table~\ref{tab:base-128-256-test}, the proposed \textbf{Sentinel+Tag} ends up outperforming all previous formats.
In fact, the \textbf{Sentinel+Tag} outperforms all the other basic approaches in 6 out of the 7 benchmarks and is significantly better when averaging out the results across all runs.

These results also show that while the \textbf{Tag Only} approach is quite competitive, using either the sentinels or input tokens leads to better results.
Additionally, we also find that the \textbf{Input + Tag} is actually marginally better than the more commonly used Tagged Spans approach (though this gap is too narrow to be statistically significant).

 \begin{table*}[h]
\centering
 \resizebox{\textwidth}{!}{
 \begin{tabular}{|l|cc|cc||cc|cc||cc||cc|cc|}
\hline
\multirow{2}{*}{Dataset} &  \multicolumn{2}{c|}{\textbf{Tag Only}} & \multicolumn{2}{c||}{\textbf{Tag Only (SI)}} & \multicolumn{2}{c|}{\textbf{Input + Tag}} & \multicolumn{2}{c||}{\textbf{Input + Tag (SI)}} & \multicolumn{2}{c||}{\textbf{Tagged Spans}} & \multicolumn{2}{c|}{\textbf{Sentinel + Tag}} & \multicolumn{2}{c|}{\textbf{Sentinel + Tag (SI)}}
 \\
& \textbf{Perfect} & \textbf{Mic F1} & \textbf{Perfect} & \textbf{Mic F1} & \textbf{Perfect} & \textbf{Mic F1} & \textbf{Perfect} & \textbf{Mic F1} & \textbf{Perfect} & \textbf{Mic F1} & \textbf{Perfect} & \textbf{Mic F1} & \textbf{Perfect} & \textbf{Mic F1}
 \\
\hline 

\textbf{mATIS} & 36.71 & 52.79 & 36.97 & 53.30 & 38.96 & 62.62 & 39.47 & 63.42 & 35.58 & 58.55 & \textbf{43.62}\textsuperscript{\textdagger} & \textbf{68.54}\textsuperscript{\textdagger} & 42.49\textsuperscript{\textdagger} & 66.67\textsuperscript{\textdagger}
\\ 
 (3 langs)  & $\pm$ 3.31 & $\pm$ 4.22 & $\pm$ 1.09 & $\pm$ 1.79 & $\pm$ 1.45 & $\pm$ 2.14 & $\pm$ 1.22 & $\pm$ 1.38 & $\pm$ 1.10 & $\pm$ 2.19 & $\pm$ 1.99 & $\pm$ 1.89 & $\pm$ 2.15 & $\pm$ 2.31
 \\

 \hline 

\textbf{mTOP} & 43.41 & 51.48 & 44.98 & 53.04 & 45.60 & 53.68 & 46.57 & 54.81 & 44.00 & 54.53 & 57.14\textsuperscript{\textdagger} & 69.65\textsuperscript{\textdagger} & \textbf{59.31}\textsuperscript{\textdagger} & \textbf{71.22}\textsuperscript{\textdagger}
\\ 
(6 langs) & $\pm$ 1.02 & $\pm$ 1.06 & $\pm$ 0.66 & $\pm$ 0.73 & $\pm$ 1.01 & $\pm$ 1.02 & $\pm$ 0.35 & $\pm$ 0.26 & $\pm$ 0.91 & $\pm$ 0.89 & $\pm$ 1.06 & $\pm$ 1.08 & $\pm$ 0.98 & $\pm$ 1.34
 \\

 \hline 
 
\textbf{mTOD} & 66.42 & 74.27 & 67.78 & 75.42 & 65.11 & 73.75 & 65.55 & 74.74 & 58.73 & 68.71 & 66.92 & 76.42\textsuperscript{\textdagger} & \textbf{70.07} & \textbf{78.51}\textsuperscript{\textdagger}
\\ 
(3 langs) & $\pm$ 1.41 & $\pm$ 1.51 & $\pm$ 1.08 & $\pm$ 0.76 & $\pm$ 1.18 & $\pm$ 1.01 & $\pm$ 0.55 & $\pm$ 0.42 & $\pm$ 1.84 & $\pm$ 1.56 & $\pm$ 1.49 & $\pm$ 0.94 & $\pm$ 1.80 & $\pm$ 1.11
 \\
 \hline

\textbf{Average} & 47.49 & 57.50 & 48.68 & 58.70 & 48.82 & 60.93 & 49.54 & 61.94 & 45.58 & 59.08 & 56.20\textsuperscript{\textdagger} & 71.07\textsuperscript{\textdagger} & \textbf{57.79}\textsuperscript{\textdagger} & \textbf{71.90}\textsuperscript{\textdagger}
\\ 
  & $\pm$ 1.09 & $\pm$ 1.31 & $\pm$ 0.39 & $\pm$ 0.81 & $\pm$ 0.37 & $\pm$ 0.46 & $\pm$ 0.46 & $\pm$ 0.33 & $\pm$ 0.57 & $\pm$ 0.82 & $\pm$ 0.95 & $\pm$ 0.79 & $\pm$ 1.38 & $\pm$ 1.40
 \\
 \hline
 \textbf{Average} & 34.17 & 45.75 & 35.51 & 47.13 & 35.58 & 49.89 & 36.47 & 51.18 & 31.29 & 47.41 & 45.20\textsuperscript{\textdagger} & 63.24\textsuperscript{\textdagger} & \textbf{47.11}\textsuperscript{\textdagger} & \textbf{64.25}\textsuperscript{\textdagger}
\\ 
 \textbf{(non-en)} & $\pm$ 1.27 & $\pm$ 1.65 & $\pm$ 0.41 & $\pm$ 0.98 & $\pm$ 0.43 & $\pm$ 0.58 & $\pm$ 0.63 & $\pm$ 0.45 & $\pm$ 0.64 & $\pm$ 1.03 & $\pm$ 1.23 & $\pm$ 1.02 & $\pm$ 1.73 & $\pm$ 1.80
 \\ \hline
 \end{tabular}
 }

\vspace{-0.05in}
 \caption{Multilingual Results in the \textbf{Zero-shot} setting (mean and standard deviation over 3 runs). Detailed (per-language) results can be found in Table~\ref{tab:multilingual-results-zs-full} in the Appendix.  \textsuperscript{\textdagger} is 98\% significance against \textbf{all} non-sentinel methods.}
 \vspace{-0.05in}
   \label{tab:multilingual-results-zs}
 \end{table*}

 \begin{table*}[!h]
\centering
 \resizebox{\textwidth}{!}{
 \begin{tabular}{|l|cc|cc||cc|cc||cc||cc|cc|}
\hline
\multirow{2}{*}{Dataset} &  \multicolumn{2}{c|}{\textbf{Tag Only}} & \multicolumn{2}{c||}{\textbf{Tag Only (SI)}} & \multicolumn{2}{c|}{\textbf{Input + Tag}} & \multicolumn{2}{c||}{\textbf{Input + Tag (SI)}} & \multicolumn{2}{c||}{\textbf{Tagged Spans}} & \multicolumn{2}{c|}{\textbf{Sentinel + Tag}} & \multicolumn{2}{c|}{\textbf{Sentinel + Tag (SI)}}
 \\
& \textbf{Perfect} & \textbf{Mic F1} & \textbf{Perfect} & \textbf{Mic F1} & \textbf{Perfect} & \textbf{Mic F1} & \textbf{Perfect} & \textbf{Mic F1} & \textbf{Perfect} & \textbf{Mic F1} & \textbf{Perfect} & \textbf{Mic F1} & \textbf{Perfect} & \textbf{Mic F1}
 \\
\hline 

\textbf{mATIS} & 68.89 & 82.90 & 70.13 & 83.94 & 71.03 & 87.22 & 70.38 & 86.69 & 69.14 & 86.53 & 73.26\textsuperscript{\textdagger} & \textbf{88.37}\textsuperscript{\textdagger} & \textbf{73.43}\textsuperscript{\textdagger} & 88.26\textsuperscript{\textdagger}
\\ 
 (3 langs)  & $\pm$ 1.09 & $\pm$ 0.74 & $\pm$ 0.46 & $\pm$ 0.50 & $\pm$ 0.47 & $\pm$ 0.39 & $\pm$ 0.54 & $\pm$ 0.26 & $\pm$ 0.21 & $\pm$ 0.23 & $\pm$ 1.17 & $\pm$ 0.37 & $\pm$ 0.74 & $\pm$ 0.53
 \\

 \hline 

\textbf{mTOP} & 69.43 & 77.31 & 73.40 & 81.00 & 75.21 & 83.10 & 76.22 & 83.90 & 73.12 & 82.10 & \textbf{81.23}\textsuperscript{\textdagger} & \textbf{88.26}\textsuperscript{\textdagger} & 80.74\textsuperscript{\textdagger} & 87.85\textsuperscript{\textdagger}
\\ 
(6 langs) & $\pm$ 0.64 & $\pm$ 0.58 & $\pm$ 0.20 & $\pm$ 0.12 & $\pm$ 0.31 & $\pm$ 0.26 & $\pm$ 0.34 & $\pm$ 0.24 & $\pm$ 0.37 & $\pm$ 0.33 & $\pm$ 0.23 & $\pm$ 0.16 & $\pm$ 0.57 & $\pm$ 0.42
 \\ 

 \hline 
 
\textbf{mTOD} & 88.81 & 92.70 & 89.58 & 93.38 & 89.61 & 93.53 & 89.63 & 93.50 & 89.19 & 93.30 & \textbf{89.97}\textsuperscript{\textdagger} & \textbf{93.72} & 89.92\textsuperscript{\textdagger} & 93.64
\\ 
(3 langs) & $\pm$ 0.14 & $\pm$ 0.09 & $\pm$ 0.26 & $\pm$ 0.18 & $\pm$ 0.22 & $\pm$ 0.17 & $\pm$ 0.17 & $\pm$ 0.10 & $\pm$ 0.22 & $\pm$ 0.15 & $\pm$ 0.08 & $\pm$ 0.07 & $\pm$ 0.12 & $\pm$ 0.12
 \\

 \hline

\textbf{Average} & 74.14 & 82.55 & 76.63 & 84.83 & 77.76 & 86.74 & 78.11 & 87.00 & 76.15 & 86.01 & \textbf{81.42}\textsuperscript{\textdagger} & \textbf{89.65}\textsuperscript{\textdagger} & 81.21\textsuperscript{\textdagger} & 89.40\textsuperscript{\textdagger}
\\ 
 & $\pm$ 0.18 & $\pm$ 0.13 & $\pm$ 0.08 & $\pm$ 0.05 & $\pm$ 0.09 & $\pm$ 0.15 & $\pm$ 0.25 & $\pm$ 0.14 & $\pm$ 0.12 & $\pm$ 0.16 & $\pm$ 0.38 & $\pm$ 0.14 & $\pm$ 0.10 & $\pm$ 0.10
 \\ \hline

\textbf{Average} & 70.37 & 79.56 & 73.22 & 82.25 & 74.60 & 84.56 & 74.89 & 84.81 & 72.60 & 83.65 & \textbf{78.94}\textsuperscript{\textdagger} & \textbf{88.08}\textsuperscript{\textdagger} & 78.78\textsuperscript{\textdagger} & 87.82\textsuperscript{\textdagger}
\\ 
\textbf{(non-en)} & $\pm$ 0.21 & $\pm$ 0.20 & $\pm$ 0.04 & $\pm$ 0.05 & $\pm$ 0.14 & $\pm$ 0.21 & $\pm$ 0.32 & $\pm$ 0.17 & $\pm$ 0.14 & $\pm$ 0.20 & $\pm$ 0.47 & $\pm$ 0.17 & $\pm$ 0.10 & $\pm$ 0.10
 \\ \hline
 \end{tabular}
 }
 \vspace{-0.05in}
  \caption{Multilingual Results in the regular \textbf{joint multilingual} setting (mean and stdev over 3 runs). Detailed (per-language) results can be found in Appendix (Table~\ref{tab:multilingual-results-full}).\textsuperscript{\textdagger} is 99\% significance against \textbf{all} non-sentinel methods.}
  \vspace{-0.1in}
   \label{tab:multilingual-results}
 \end{table*}

\subsection{Do variants \& simplifications work?}
\label{sec:exp-results1-variants}
We next tried to understand performance on making tweaks to the formats as discussed in Sec~\ref{sec:format-variants}.
The last 3 columns of Table~\ref{tab:base-128-256-test}, includes results for variants that simplify the Inside tag to just "\textit{I}.
Results for the mBERT baseline with the same simplification are also provided.
Overall the \textbf{Sentinel + Tag (SI)} resulted in the aggregate best performance across all formats on the 7 English benchmarks. 
The small gap between SI and non-SI variants makes sense given the autoregressive nature of the decoder (that can generate the full Inside tags by attending to any opened Begin tag that need to be continued).
We should note that the \textbf{SI} variants have one significiant advantage.
As seen in Table~\ref{tab:seq-len-stats}, these formats have significantly shorter target sequences across all datasets, resulting in faster training and inference.

Another simplification we evaluated involved the \emph{Outside} tags.
Interestingly, as seen in Table~\ref{tab:base-128-256-test-so}, results worsen when using this simplification -- for all formats -- in nearly all settings.
This indicates that these Seq2Seq models find it easier to produce a consistent prediction for all tokens rather than having to skip tokens.
More details and analysis for this simplification can be found in Appendix~\ref{sec:appendix-format-variant}.


%% file: 7-multilinguality.tex
\section{Multilingual Capabilities}
\label{sec:exp-multilinguality}
\vspace{-0.1in}

A major benefit of large LMs is their generalization capabilities in the multilingual setting.
In this section, we evaluated the different formats in the more challenging multilingual setting.
Specifically we ran experiments on 3 benchmarks (spanning 7 different languages) in two different settings:
\begin{itemize}[noitemsep,topsep=0pt]
    \item \textbf{Joint Multilingual}: Models are trained by combining data for all languages. The weights for all languages are equal \emph{i.e.,} the models see roughly the same number of examples in all languages. For simplicity, checkpoint selections happens per-language using that language's validation set (though we did not find this choice to affect results significantly).
    \item \textbf{Zero-shot}: Here we train the model using only English data, and evaluate on all languages. This is the more challenging setting for evaluating multilingual generalization.
\end{itemize}
As seen in the results for the zero-shot setting in Table~\ref{tab:multilingual-results-zs}, the findings from the previous section are only further emphasized with the \textbf{Sentinel+Tag} approaches significantly outperforming all the other approaches.
For instance, when looking at the \textit{Perfect} metrics averaged across all 9 non-English test sets (2 mATIS, 5 mTOP, 2 mTOD), we find \textbf{Sentinel+Tag (SI)} format greatly outperforms all non-sentinel formats with the next best alternative (\textbf{Input+Tag}) being more than \textbf{10.5\%} absolute worse (\textbf{47.1\%} vs \textbf{36.5\%}, a \textbf{30\%} relative increase in accuracy).
Furthermore it is nearly \textbf{16\%} better (a \textbf{47\%} increase) than the current standard \emph{i.e.,} -- \textbf{Tagged Spans} (\textbf{31.3} vs \textbf{47.1}).

\begin{table}[!t]
\centering
 \resizebox{\textwidth / 2}{!}{
 \begin{tabular}{|l|cc|cc|cc|}
\hline
\multirow{2}{*}{Dataset} &  \multicolumn{2}{c|}{\textbf{Extractive TS}} & \multicolumn{2}{c|}{\textbf{Extractive S+T}} & \multicolumn{2}{c|}{\textbf{Extractive S+T(S)}}
 \\
& \textbf{Perfect} & \textbf{Mic F1} & \textbf{Perfect} & \textbf{Mic F1} & \textbf{Perfect} & \textbf{Mic F1} 
 \\
\hline 

\textbf{mATIS} & 35.17 & 57.56 & 37.62 & 64.07\textsuperscript{\textdagger} & \textbf{42.17}\textsuperscript{\textdagger} & \textbf{67.48}\textsuperscript{\textdagger}
\\ 
 (3 langs)  & $\pm$ 0.38 & $\pm$ 0.84 & $\pm$ 2.87 & $\pm$ 2.13 & $\pm$ 0.91 & $\pm$ 0.66
 \\

 \hline 

\textbf{mTOP} & 65.43 & 48.39 & 80.14\textsuperscript{\textdagger} & 67.37\textsuperscript{\textdagger} & \textbf{80.81}\textsuperscript{\textdagger} & \textbf{68.57}\textsuperscript{\textdagger}
\\ 
(6 langs) & $\pm$ 1.16 & $\pm$ 1.79 & $\pm$ 1.00 & $\pm$ 1.99 & $\pm$ 1.01 & $\pm$ 1.18
 \\

 \hline 
 
\textbf{mTOD} & 52.85 & 66.46 & 60.11\textsuperscript{\textdagger} & 72.74\textsuperscript{\textdagger} & \textbf{61.47}\textsuperscript{\textdagger} & \textbf{73.66}\textsuperscript{\textdagger}
\\ 
(3 langs) & $\pm$ 2.60 & $\pm$ 2.47 & $\pm$ 1.05 & $\pm$ 0.44 & $\pm$ 0.39 & $\pm$ 0.26
 \\
 \hline

\textbf{Average} & 54.72 & 55.20 & 64.50\textsuperscript{\textdagger} & 67.89\textsuperscript{\textdagger} & \textbf{66.31}\textsuperscript{\textdagger} & \textbf{69.57}\textsuperscript{\textdagger}
\\ 
  & $\pm$ 0.17 & $\pm$ 0.06 & $\pm$ 1.48 & $\pm$ 1.42 & $\pm$ 0.83 & $\pm$ 0.82
 \\
 \hline
 \textbf{Average} & 43.07 & 42.46 & 56.34\textsuperscript{\textdagger} & 59.60\textsuperscript{\textdagger} & \textbf{58.55}\textsuperscript{\textdagger} & \textbf{61.64}\textsuperscript{\textdagger}
\\ 
 \textbf{(non-en)} & $\pm$ 0.20 & $\pm$ 0.09 & $\pm$ 1.74 & $\pm$ 1.69 & $\pm$ 1.05 & $\pm$ 1.04
 \\ \hline
 \end{tabular}
 }

\vspace{-0.05in}
 \caption{\textbf{Zero-shot} results (mean and stdev over 2 runs) for \textbf{extractive} methods. Per-language results reported in  Table~\ref{tab:multilingual-results-zs-extractive-full}.\textsuperscript{\textdagger} indicates 99\% significance.}
 \vspace{-0.2in}
   \label{tab:multilingual-results-zs-extractive}
 \end{table}

These gains are all the more noteworthy when compared to the performance of the \textbf{Tag Only} (no sentinels) and \textbf{Input+Tag} (input token instead of sentinel) approaches.
While those two formats have very similar performance -- with \textbf{Input+Tag} being slightly better than \textbf{Tag Only} --  the addition of the sentinel tokens drastically improves performance.
Specifically, the language-agnostic and input-agnostic nature of these sentinel tokens enables the language-agnosticity of the decoder leading to improved generalization across languages despite the training data being only English.

As seen in Table~\ref{tab:multilingual-results}, these trends largely continue in the joint multilingual training setting as well.
Despite the use of training data from all languages, the \textbf{Sentinel+Tag} formats outperform all non-sentinel formats by between \textbf{4 and 9 pp} on the Perfect metric.
We again find a large gap (\textbf{6.3 pp}) to the current \textbf{Tagged Spans} approach (a \textbf{8.7\%} increase from \textbf{72.6} to \textbf{78.9}).

A few additional observations: (a)  As on English benchmarks, the Inside simplification slightly improved performance  -- though the gaps are within error margins. (b) As seen in Table ~\ref{tab:base-128-256-test-so}, the Outside simplification tends to hurt zeroshot performance significantly for all techniques  (c) \textbf{Tagged Spans} -- the current standard in published works -- is significantly worse than other formats in both settings and especially in the zero-shot setting.

%% file: X-extractive.tex
 \begin{table}[!t]
\centering
 \resizebox{\textwidth / 2}{!}{
 \begin{tabular}{|l|cc|cc|cc|}
\hline
\multirow{2}{*}{Dataset} &  \multicolumn{2}{c|}{\textbf{Extractive TS}} & \multicolumn{2}{c|}{\textbf{Extractive S+T}} & \multicolumn{2}{c|}{\textbf{Extractive S+T(S)}}
 \\
& \textbf{Perfect} & \textbf{Mic F1} & \textbf{Perfect} & \textbf{Mic F1} & \textbf{Perfect} & \textbf{Mic F1} 
 \\
\hline 

\textbf{mATIS} & 70.01 & 86.66 & 72.06 & 87.75 & \textbf{72.94}\textsuperscript{\textdagger} & \textbf{88.09}\textsuperscript{\textdagger}
\\ 
 (3 langs)  & $\pm$ 1.20 & $\pm$ 0.48 & $\pm$ 1.08 & $\pm$ 0.55 & $\pm$ 0.34 & $\pm$ 0.31
 \\

 \hline 

\textbf{mTOP} & 87.78 & 77.17 & 90.57\textsuperscript{\textdagger} & 84.15\textsuperscript{\textdagger} & \textbf{91.89}\textsuperscript{\textdagger} & \textbf{86.02}\textsuperscript{\textdagger}
\\ 
(6 langs) & $\pm$ 0.30 & $\pm$ 0.51 & $\pm$ 0.87 & $\pm$ 0.17 & $\pm$ 0.00 & $\pm$ 0.80
 \\

 \hline 
 
\textbf{mTOD} & \textbf{89.02} & \textbf{93.32} & 87.48 & 92.36 & 88.58 & 92.88\textsuperscript{\textdagger}
\\ 
(3 langs) & $\pm$ 0.16 & $\pm$ 0.03 & $\pm$ 0.70 & $\pm$ 0.40 & $\pm$ 0.30 & $\pm$ 0.24
 \\
 \hline

\textbf{Average} & 83.65 & 83.58 & 85.17\textsuperscript{\textdagger} & 87.10\textsuperscript{\textdagger} & \textbf{86.33}\textsuperscript{\textdagger} & \textbf{88.26}\textsuperscript{\textdagger}
\\ 
  & $\pm$ 0.49 & $\pm$ 0.38 & $\pm$ 0.53 & $\pm$ 0.12 & $\pm$ 0.16 & $\pm$ 0.54
 \\
 \hline
 \textbf{Average} & 82.25 & 80.72 & 84.11\textsuperscript{\textdagger} & 85.24\textsuperscript{\textdagger} & \textbf{85.04}\textsuperscript{\textdagger} & \textbf{86.40}\textsuperscript{\textdagger}
\\ 
 \textbf{(non-en)} & $\pm$ 0.57 & $\pm$ 0.43 & $\pm$ 0.42 & $\pm$ 0.00 & $\pm$ 0.16 & $\pm$ 0.71
 \\ \hline
 \end{tabular}
 }

\vspace{-0.05in}
 \caption{\textbf{Joint multilingual} results (mean and stdev over 2 runs) for \textbf{extractive} methods. Per-language results reported in  Table~\ref{tab:multilingual-results-joint-extractive-full}.\textsuperscript{\textdagger} indicates 99\% significance.}
 \vspace{-0.15in}
   \label{tab:multilingual-results-joint-extractive}
 \end{table}

\begin{table*}[ht]
\centering
 \resizebox{\textwidth}{!}{
 \begin{tabular}{|l|cccc||cccc||cccc||cccc|}
\hline
\multirow{2}{*}{Size} & \multicolumn{4}{c||}{\textbf{Average (all) [Zero-Shot]}} & \multicolumn{4}{c||}{\textbf{Average (non-en) [Zero-Shot]}} & \multicolumn{4}{c||}{\textbf{Average (all) [Joint]}} & \multicolumn{4}{c|}{\textbf{Average (non-en) [Joint]}}
 \\
& \textbf{T-O} & \textbf{I+T} & \textbf{TS} & \textbf{S+T} & \textbf{T-O} & \textbf{I+T} & \textbf{TS} & \textbf{S+T} & \textbf{T-O} & \textbf{I+T} & \textbf{TS} & \textbf{S+T} & \textbf{T-O} & \textbf{I+T} & \textbf{TS} & \textbf{S+T}
 \\
\hline 

Small    
& 44.57 & 44.96 & 42.78 & \textbf{46.61}\textsuperscript{\textdagger}
& 30.28 & 30.73 & 27.96 & \textbf{32.49}\textsuperscript{\textdagger}
& 74.53 & 75.52 & 73.64 & \textbf{78.55}\textsuperscript{\textdagger}
& 71.07 & 72.25 & 70.14 & \textbf{75.76}\textsuperscript{\textdagger}
\\
(300M)
& $\pm$ 0.03 & $\pm$ 0.00 & $\pm$ 0.05 & $\pm$ 0.01
& $\pm$ 0.07 & $\pm$ 0.02 & $\pm$ 0.15 & $\pm$ 0.01
& $\pm$ 0.06 & $\pm$ 0.13 & $\pm$ 0.29 & $\pm$ 0.06
& $\pm$ 0.11 & $\pm$ 0.20 & $\pm$ 0.31 & $\pm$ 0.09

\\
\hline

Base
& 48.68 & 49.54 & 45.58 & \textbf{57.79}\textsuperscript{\textdagger}
& 35.51 & 36.47 & 31.29 & \textbf{47.11}\textsuperscript{\textdagger}
& 76.63 & 78.11 & 76.15 & \textbf{81.21}\textsuperscript{\textdagger}
& 73.22 & 74.89 & 72.60 & \textbf{78.78}\textsuperscript{\textdagger}
\\ 
(580M)
& $\pm$ 0.39 & $\pm$ 0.46 & $\pm$ 0.57 & $\pm$ 1.38
& $\pm$ 0.41 & $\pm$ 0.63 & $\pm$ 0.64 & $\pm$ 1.73
& $\pm$ 0.08 & $\pm$ 0.25 & $\pm$ 0.12 & $\pm$ 0.10
& $\pm$ 0.04 & $\pm$ 0.32 & $\pm$ 0.14 & $\pm$ 0.10
 \\ 
\hline

Large
& 51.26 & 50.27 & 46.15 & \textbf{59.80}\textsuperscript{\textdagger}
& 38.96 & 37.12 & 31.76 & \textbf{49.74}\textsuperscript{\textdagger}
& 79.24 & 80.03 & 78.51 & \textbf{81.37}\textsuperscript{\textdagger}
& 76.24 & 77.05 & 75.13 & \textbf{78.89}\textsuperscript{\textdagger}
\\
(1.2B)
& $\pm$ 0.55 & $\pm$ 0.62 & $\pm$ 0.28 & $\pm$ 0.89
& $\pm$ 0.68 & $\pm$ 0.79 & $\pm$ 0.41 & $\pm$ 1.13
& $\pm$ 0.11 & $\pm$ 0.19 & $\pm$ 0.24 & $\pm$ 0.10
& $\pm$ 0.15 & $\pm$ 0.19 & $\pm$ 0.31 & $\pm$ 0.14
\\

\hline
XL
& 54.72 & 59.01 & 60.27 & \textbf{64.47}\textsuperscript{\textdagger}
& 42.86 & 48.49 & 50.17 & \textbf{55.70}\textsuperscript{\textdagger}
& 82.44 & 82.63 & 81.73 & \textbf{84.04}\textsuperscript{\textdagger}
& 79.73 & 79.97 & 78.80 & \textbf{81.73}\textsuperscript{\textdagger}

\\
(3.7B)
& $\pm$ 0.32 & $\pm$ 0.22 & $\pm$ 0.84 & $\pm$ 0.08
& $\pm$ 0.39 & $\pm$ 0.30 & $\pm$ 1.18 & $\pm$ 0.12
& $\pm$ 0.21 & $\pm$ 0.07 & $\pm$ 0.04 & $\pm$ 0.10
& $\pm$ 0.30 & $\pm$ 0.12 & $\pm$ 0.05 & $\pm$ 0.10
\\
\hline

XXL
& 53.27 & 57.32 & 59.97 & \textbf{65.55}
& 40.95 & 46.27 & 49.84 & \textbf{57.36}
& 82.13 & 82.59 & 82.20 & \textbf{83.86}
& 79.35 & 79.91 & 79.34 & \textbf{81.62}

\\
\hline

 \hline
 \end{tabular}
 }

\vspace{-0.05in}
  \caption{Impact of model size (\# of params in parentheses) on performance of formats. Averaged Perfect metric scores are reported over the same 3 benchmarks (12 test sets) as Table~\ref{tab:multilingual-results-zs}, on both zero-shot and joint settings. The 4 methods compared are \textbf{T-O}: Tag Only (SI), \textbf{I+T}: Input + Tag (SI), \textbf{TS}: Tagged Spans and \textbf{S+T}: Sentinel + Tag (SI). Base results are averaged over 3 runs. XL, Large and Small were averaged over 2 runs. Due to compute limits, XXL (13B params) was run once for 2k steps (since trial runs plateaued there).\textsuperscript{\textdagger} indicates 99\% significance.}
   \label{tab:multilingual-model-size}
 \end{table*}

\begin{table*}[h]
\centering
 \resizebox{\textwidth}{!}{
 \begin{tabular}{|l|cccc||cccc||cccc||cccc|}
\hline
\multirow{2}{*}{Size} & \multicolumn{4}{c||}{\textbf{Average (all) [Zero-Shot]}} & \multicolumn{4}{c||}{\textbf{Average (non-en) [Zero-Shot]}} & \multicolumn{4}{c||}{\textbf{Average (all) [Joint]}} & \multicolumn{4}{c|}{\textbf{Average (non-en) [Joint]}}
 \\
& \textbf{T-O} & \textbf{I+T} & \textbf{TS} & \textbf{S+T} & \textbf{T-O} & \textbf{I+T} & \textbf{TS} & \textbf{S+T} & \textbf{T-O} & \textbf{I+T} & \textbf{TS} & \textbf{S+T} & \textbf{T-O} & \textbf{I+T} & \textbf{TS} & \textbf{S+T}
 \\
\hline 

Small    
& 29.92 & 33.88 (2.5) & 32.14 & \textbf{0.03}
& 39.14 & 44.99(3.32)  & 42.61 & \textbf{0.03}
& 8.90 & 5.81 (0.88) & 6.19 & \textbf{0.00}
& 10.82 & 7.61 (1.18) & 8.10 & \textbf{0.00}

\\

\hline

Base
& 29.68 & 40.60 (4.43) & 43.22 & \textbf{0.02}
& 38.87 & 53.89 (5.91) & 57.38 & \textbf{0.02}
& 8.53 & 5.42 (0.9) & 6.10 & \textbf{0.01}
& 10.58 & 7.06 (1.15) & 7.87 & \textbf{0.01}

\\ 

\hline

Large

& 30.60 & 40.14  (1.9) & 47.78 & \textbf{0.07}
& 39.67 & 53.37 (2.54) & 63.48 & \textbf{0.09}
& 8.68 & 5.04 (0.87) & 5.81 & \textbf{0.00}
& 10.71 & 6.52 (1.21) & 7.53 & \textbf{0.00}

\\



\hline
XL
& 29.92 & 24.76 (1.55) & 20.39 & \textbf{0.11}
& 39.64 & 32.96 (2.08) & 27.15 & \textbf{0.14}
& 3.20 & 3.81 (0.66) & 4.10 & \textbf{0.01}
& 4.12 & 5.04 (0.95) & 5.42 & \textbf{0.01}



\\
\hline

XXL
& 32.81 & 32.64 (2.16) & 21.35 & \textbf{0.68}
& 43.41 & 43.46 (2.89) & 28.34 & \textbf{0.88}
& 4.91 & 3.71 (0.56) & 3.79 & \textbf{0.00}
& 6.38 & 4.90 (0.75)& 4.99 & \textbf{0.00}

\\
\hline

 \hline
 \end{tabular}
 }

 \vspace{-0.05in}
  \caption{Percentage of examples with hallucinations observed for the different formats / models from Table~\ref{tab:multilingual-model-size}. For \textbf{Input+ Tag} we also report (in parenthesis) the \% of examples with hallucinations in correctly tagged spans.}
  \vspace{-0.15in}
   \label{tab:multilingual-hallucination}
 \end{table*}

\section{Performance of Extractive methods}
\label{sec:exp-extractive}
\vspace{-0.05in}

As discussed in Sec~\ref{sec:format-extractive}, certain applications may prefer extractive formats. To verify if our findings extend to these settings we also evaluated these extractive formats on the same datasets. Tables~\ref{tab:multilingual-results-zs-extractive} (zeroshot) and~\ref{tab:multilingual-results-joint-extractive} (joint multilingual) summarize the results. A few key observations:
\begin{itemize}[noitemsep,topsep=0pt,leftmargin=*]
    \item The extractive sentinel-based approaches continue the trend of significantly outperform existing extractive approaches across all metrics.
    \item When comparing Tables~\ref{tab:multilingual-results-zs} vs~\ref{tab:multilingual-results-zs-extractive} and Tables~\ref{tab:multilingual-results} vs~\ref{tab:multilingual-results-joint-extractive}, we find that all the Extractive formats improve on the \emph{Perfect} metric as opposed to formats which output labels for the entire input. On further analysis we find this is due to the simpler (and shorter length) prediction task in this case.
    \item However, when we look at the \emph{MicroF1} metrics, we find that extractive models are \textbf{worse} at \emph{extracting} the correct labeled spans than models that simply label every input token. This performance drop is particularly significant for the existing (TaggedSpans) approach (3-5pp) -- vs. the drop in sentinel models (1.2-2.5pp).
\end{itemize}
These results may have wider implications on how we should model extractive tasks and the formats used there. However we leave this to future work.

%% file: X-robustness.tex
\section{Robustness \& Efficiency}
\label{sec:exp-robustness}

Beyond performance, efficiency and robustness of a model are equally important factors in practice. We focus on measuring these in this section.

\vspace{-0.05in}
\subsection{Understanding effect of model size}
\label{sec:exp-robustness-model-size}
\vspace{-0.05in}

While experiments and results so far used \textit{Base} sized models (to maximize experimentation given compute), we would like to understand how robust our findings are as we vary model sizes.
As seen in Table~\ref{tab:multilingual-model-size} (and Table~\ref{tab:multilingual-model-size-f1}),  the previously observed trends are all consistently repeated across all model sizes.
In particular, we still find the \textbf{Sentinel+Tag} models performing the best across all the datasets for both the zero-shot and joint settings.
Notably, zero-shot performance improves significantly for all datasets as model size increases.
Consequently, the larger models have a smaller gap between zero-shot and joint settings.
Put together, the performance of the \textbf{Sentinel+Tag} format is robust across all model sizes, datasets, and experiment settings.

\begin{table}[!t]
\centering
 \resizebox{\textwidth / 2}{!}{
 \begin{tabular}{l|ccc||ccc}
\multirow{2}{*}{Dataset} &  \multicolumn{3}{c||}{\textbf{Zeroshot}} & \multicolumn{3}{c}{\textbf{Joint}}
 \\

& \textbf{E TS} & \textbf{E S+T} & \textbf{E S+T(S)} & \textbf{E TS} & \textbf{E S+T} & \textbf{E S+T(S)}
 \\
\hline 

\textbf{mATIS} (3 langs) & 32.65 & 0.41 & \textbf{0.17} & 4.09 & 0.07 & \textbf{0.06}
\\ 

 \hline 

\textbf{mTOP} (6 langs) & 17.68 & 0.18 & \textbf{0.03} & 0.56 & 0.01 & \textbf{0.00}
\\ 
 \hline 
 
\textbf{mTOD} (3 langs) & 24.59 & \textbf{0.12} & \textbf{0.12} & 0.31 & 0.11 & \textbf{0.07}
\\ 
 \hline

\textbf{Average} & 23.15 & 0.22 & \textbf{0.09} & 1.38 & 0.05 & \textbf{0.03}
\\ 
 \hline
\textbf{Average (non-en)} & 30.75 & 0.25 & \textbf{0.10} & 1.74 & 0.05 & \textbf{0.03}
\\ 
 \end{tabular}
 }

\vspace{-0.05in}
 \caption{\% of examples with hallucinations for different \textbf{extractive} formats. Per-dataset results in Table~\ref{tab:multilingual-hallucinations-extractive-full}.}
 \vspace{-0.15in}
   \label{tab:multilingual-hallucinations-extractive}
 \end{table}

\subsection{Understanding effect of hallucination}
\label{sec:hallucinations}

One of our key motivations for exploring alternative formats,  is the prevalence of hallucination.
Hallucination is one of the most notorious problems plaguing Seq2Seq models and hurting practical adoption.
In particular, we found that models trained using the \textbf{Tagged Spans} or \textbf{Input + Tag} formats, often resulted in outputs containing words that not present in the input (even on English examples).
Even our previous metrics for the \textbf{Input + Tag} formats were generous as they glossed over some hallucinations by comparing (tagged) token indices rather than the actual texts.

We measured hallucinations for the different formats in a simple, straightforward manner.
In particular, for formats that repeated the input text in the output (\textbf{Tagged Spans} and \textbf{Input + Tag}) we measured how often the generated sentence does not match the input (\emph{i.e.,} one of the words are either deleted, inserted or modified).
For formats which do not repeat the input, we counted the number of tokens and checked if they match -- for \textbf{Tag Only} this was the number of tags produced in the output, whereas for \textbf{Sentinel + Tag} we counted the generated sentinel tokens.

Table~\ref{tab:multilingual-hallucination} shows how the rate of hallucination across different model sizes and formats.
As noted earlier, we find that hallucination is highly prevalent in the \textbf{Tagged Spans} and \textbf{Input + Tag} formats due to having to repeat the input -- which often leads to errors.
In the zero-shot setting, we find that nearly a third of all the examples had one of more hallucinations in most models for these formats.
Furthermore, these hallucinations were present in both English as well as other languages -- though slightly less common in English.
Even in the joint multilingual setting, we found that about 1 in every 20-25 predictions had some hallucination.
Appendix~\ref{sec:hallucination-evals} contains examples of hallucinations for different models on these datasets.

\begin{table}[!t]
\centering
 \resizebox{\textwidth / 2}{!}{
\begin{tabular}{|l|ll|}
 \hline
 
\textbf{Format} & \multicolumn{2}{|c|}{\textbf{Average / 99\%-ile Seq. Len} (and variants)}
\\
  \hline
  
  \multicolumn{3}{|c|}{\textbf{Input Lengths}} \\ \hline

Original Input &  \multicolumn{2}{|l|}{14.8 / 32.6} \\ 
Sentinel Input & \multicolumn{2}{|l|}{25.2 / 55.8} \\ \hline

\multicolumn{3}{|c|}{\textbf{Target / Output Lengths}} \\ \hline

Tag Only & 23.3 / 70.4 & \textbf{With SI:} 15.5 / 34.7 \\  
Input + Tag & 57.9 / 142.2 & \textbf{With SI:} 50.0 / 109.2,  \textbf{SI+SO:} 33.8 / 82.3 \\ 
Tagged Spans & 58.3 / 123.4 & \textbf{With SO:} 31.3 / 67.9 \\  
Sentinel + Tag & 33.0 / 90.1 & \textbf{With SI:} 25.2 / 54.8,  \textbf{SI+SO:} 19.8 / 45.8 \\ 
\hline
\end{tabular}
 }

\vspace{-0.05in}
 \caption{Sequence length (\emph{i.e.,} the number SentencePiece tokens) statistics (averaged) across datasets for the different formats studied (Full stats in Table~\ref{tab:seq-len-stats-full}).}
 \vspace{-0.2in}
 \label{tab:seq-len-stats}
 \end{table}

On the other hand, the \textbf{Sentinel + Tag} based models are almost completely devoid of hallucination, with either zero or near zero hallucinated instances across all model sizes in both settings across all datasets.
This order(s) of magnitude difference in robustness to hallucination serves as clear evidence of the sentinel-based approach being more robust and practical.

\textbf{Extractive formats:} A curious reader may wonder if the shorter, more succinct extractive formats still suffer from this issue. The results can be found in Table~\ref{tab:multilingual-hallucinations-extractive}. We find similar trends here too, with existing (input-repeating) models having orders of magnitude more hallucinations than the sentinel-based formats.
Examples of these hallucinations can be found in Appendix~\ref{sec:hallucination-examples-extractive}.
Additionally, we also provide examples of other (non-hallucination) wins and losses of the sentinel approach in Appendix~\ref{sec:wins-losses-examples}.


\vspace{-0.05in}
\subsection{Efficiency and effect of sequence length}
\label{sec:exp-robustness-seq-len}


While running experiments, we observed that \textbf{Sentinel+Tag} and \textbf{Tag-Only} models were much faster at both training and especially at inference.
The key to this is the sequence lengths of these models being far shorter than those for models which need the input to be repeated like the \textbf{Input+Tag} and \textbf{Tagged Spans} approaches.
In particular, as seen in Table~\ref{tab:seq-len-stats}, the outputs for the best performing \textbf{Sentinel+Tag (SI)} format are often less than half the length of those of the current standard -- \textbf{Tagged Spans}.
While the sentinel-augmented inputs are longer (about 10 tokens on average), this difference was much smaller than the difference in the output lengths (33 tokens on average).
This trend holds across all datasets, languages and aggregating functions (average, median, $99^{th}$ percentile ..).
Given the quadratic complexity of sequence length, such efficient gains are invaluable in practical scenarios.
Consequently even though some simplifications -- like the Simplified Outside variant of the \textbf{Sentinel+Tag} -- did not improve performance, being 25-30\% the length of current formats can translate to \textbf{10x+} speedup in training and inference.

Additionally, as detailed in Appendix~\ref{sec:effect-of-seq-len} and Table~\ref{tab:multilingual-seq-len} we also verified that the new format is robust to changes in training sequence lengths.

%% file: X-discussions.tex
\section{Conclusion}
\label{sec:discussions}

In this paper, we rigorously studied different input and output sequence \emph{formats} for casting Sequence Tagging tasks as Seq2Seq problems. 
Using experiments across 7 public benchmarks, we found that the formats of the input and output sequences can significantly impact performance.
To remedy the shortcomings of the existing formats, we also introduced a new format that uses sentinel tokens.
Along with its variants, this new format proved to not only be simpler and more computationally efficient but also more accurate.
The sentinel-based formats significantly outperform all other formats when it comes to multilingual generalization, with 30+\% relative improvement in accuracy.
While current formats are plagued by hallucination in a large percentage of examples, the new format rectifies this issue and is virtually hallucination free.



%% file: limitations.tex
\newpage
\clearpage

\section{Limitations}

There are two notable limitations we would like to discuss of our work -- specifically the proposed sentinel-based approach.

\begin{enumerate}[label=\Alph*]
    \item \textbf{Need for additional preprocessing and post-processing:} The sentinel approach requires that \emph{sentinel} tokens be inserted into the input, so as to simplify decoding (\emph{i.e.,} enable the model to skip generating the original input tokens). While this leads a to host of gains as mentioned in the paper, it would be remiss of us to not point out that this approach comes with a drawback. Namely, the need for additional pre- and post-processing. Inserting the sentinels into the input required some (potentially lightweight) tokenization or lexing of the input sentence. While this may not be an issue for English and other popularly studied languages, this can be non-trivial for non-space separated languages like Thai and Lao or agglutinative languages like Finnish and German. While we tried to study the impact of this in Appendix B and found no performance loss, we should note that this is a function of the quality of the pre-processing tokenizer / lexer. If tokens are often accidentally split across tag boundaries by this pre-processing step, then this may hurt performance. Without a more detailed study on these languages (which is somewhat challenged by availability of suitable datasets in these languages), it is hard to quantify the effect this need for pre-processing may require in general.
    
    From a practical viewpoint, this pre-processing also adds an additional (albeit likely lightweight) step before the model, and potentially an additional post-processing step (joining input tokens and predicted tags) based on the output data format required by downstream steps. While we found the computation savings from the sentinel format more than enough to counteract any additional preprocessing, this is a function of application and specific system specifications and hence worth noting for practitioners.
    
    \item \textbf{Efficiency for long texts on (short) extractive tasks:} Due to the insertion of the sentinel tags we increase the sequence length of our inputs (by about 10 tokens on average in our studied datasets per Table 10). This is more than made up for by the larger decrease in output sequence length on the datasets we studied. However, for extractive tasks we can simply only output extracted spans and their associated tasks. In our running example (Table 2), this may perhaps look like: "ARTIST: Kent James, PLAYLIST: Disney". While a non-sentinel approach for such an output still suffers many of same issues as the non-sentinel formats studied in our work (namely lower performance, worse multilingual generalization, frequent hallucination ..) the efficiency gains from the sentinel approach are far less clear. In particular, when we have such an extractive task with \textbf{long} inputs (think 500+ sequence length) but very short output (\emph{i.e.,} very few extractions), then the sentinel approach would likely be slower (at training and inference) than a non-sentinel extractive output format.
    
    While this is not a problem setting we work on or focused on in this paper, we fully recognize that a non-trivial fraction of the community may have interest in such a setting. Hence we find this important to point out.
    
\end{enumerate}

%% file: 10-responsible-nlp.tex
\section{Ethics and Responsible NLP}

This paper tried to provide empirical insights into how sequence tagging NLP tasks should be handled in the Seq2Seq regime given the increased prevalence of such pretrained models.
Along the way it proposed a new \textbf{Sentinel}-based approach to the problem.
Given the nature of the work and findings, we could not think of any explicit \emph{risks} associated with this work or the new format.
Instead we could postulate about potential benefits these new \textbf{Sentinel}-based formats could bring about with wider adoption:
\begin{itemize}
    \item \textbf{Better multilingual generalization}: Our results demonstrate the potential to improve performance of common NLP tasks for low-resource languages (in addition to gains on high-resource languages).
    \item \textbf{More trustworthy NLP models}: By nearly entirely eliminating hallucinations, the new format has promise to increase fidelity of NLP models.
    \item \textbf{Reducing compute}: As discussed in Section~\ref{sec:exp-robustness-seq-len}, the new approach could potentially lead to an order of magnitude reduction in training / inference time versus the current status quo for these problems. Furthermore, given concerns surrounding massive compute models, we verified our findings (like increased multilinguality and near-zero hallucination) hold true for even the "Small" and "Base" sized pretrained mT5 models (see Tables~\ref{tab:multilingual-model-size} and \ref{tab:multilingual-hallucination}).
    \item \textbf{Possible benefits to privacy}: While not discussed in detail in the paper, one notable benefit of adding the sentinel tokens is that the output of the model no longer contains any input token. This is very amenable to privacy-preserving / privacy-focused NLP applications and potentially unlocks new opportunities for these kinds of models in more privacy sensitive settings.
\end{itemize}

%% file: appendix-all.tex

\section{Experiments with Other Format Variants}
\label{sec:appendix-format-variant}
In addition to the four primary formats and their variants with Simplified-Inside tags, we show examples of several other variants that we studied in  Table~\ref{tab:format-variant-examples}.

For example, we experimented with the Simplified-Output (SO) variant, which omits all the Outside tags. 
The results in Table~\ref{tab:base-so-results} shows that the models do not benefit from skipping any tokens or tags.
While these variants result in shorter output sequences (and therefore more efficient computation), the accuracy of these methods were lower than that of the Sentinel + Tag (and its Simplified-Inside variant) as shown in Table~\ref{tab:base-so-results}.

\begin{table}[h]
\centering
 \resizebox{\textwidth / 2}{!}{
 \begin{tabular}{|l|c|c|c|}
  \hline

\textbf{Dataset} & \textbf{Input + Tag} & \textbf{Tagged Spans} & \textbf{Sentinel +} 
\\
 & \textbf{(SI, SO)} & \textbf{(SO)} & \textbf{Tag (SI, SO)} 
  \\
  \hline

\textbf{mATIS}(en) & 88.80 & 89.03 & \textbf{90.15} 
 \\ \hline
\textbf{SNIPS} & 88.71  & 88.14  & \textbf{90.43} 
 \\ \hline
\textbf{MovieTrivia} & 37.33 & 37.63 & 35.59 
 \\ \hline
\textbf{Movies} & 70.95 & 71.61 & 71.08 
 \\ \hline
\textbf{Restaurant} & 59.50 & 60.82 & 62.80 
 \\ \hline
\textbf{mTOP}(en) & 82.65 & 84.09 & 86.34 
 \\ \hline
\textbf{mTOD}(en) & 92.50 & 92.61 & \textbf{93.36} 
 \\ \hline
\hline
\textbf{Average} & 74.35 & 74.85 & 75.68 
\\
($\Delta$ vs. non-SO) & -1.13 & -0.26 & -0.77 

 \\ \hline
 \end{tabular}
 }

 \caption{Per-dataset \textbf{Perfect} metric scores for the variant formats that modify or drop the Outside tag.}
   \label{tab:base-so-results}
 \end{table}

\section{Sentinels in complex languages}

In languages like Thai and Lao, sentences can often be composed of just one or two "\textit{words}" \emph{i.e.,} because unlike English their words are not separated by spaces.
Similarly agglutinative languages like Finnish (and to a far lesser extent German) may join multiple affixes / morphemes to form very long "\textit{words}".

This brings up the question of how we should insert sentinels for inputs in these languages, since we may be actually altering the underlying token representations.
For example, say our input was the German phrase \textit{
schweres werkzeug} (heavy tool) and assume it was tokenized as "schweres" + "werk" + "zeug".
When we create the sentinel input, we could either use: "<S> schweres <S> werk <S> zeug" or as "<S> schweres <S> werk <S>zeug" (with "<S>" representing the sentinel tokens).
The two texts would be tokenized differently by the SentencePiece model (SPM) due to the lack of a space between the sentinel and "zeug" -- since SPM treats continued tokens differently than word starts.

To understand if these models are robust to that we ran experiments on the German and Thai datasets.
As seen in Table~\ref{tab:multilingual-space-sentinel}, while there are slight differences, the models are quite robust to the tokenization and how / where the sentinels tokens are introduced -- another evidence of the robustness of the sentinel based approach.

\begin{table}[!h]
\centering
 \resizebox{\textwidth / 2}{!}{
 \begin{tabular}{|l|c|c||c|c|}
 \toprule
\multirow{2}{*}{Dataset} & \multicolumn{2}{c||}{\textbf{Space inserted}} & \multicolumn{2}{c|}{\textbf{No space}}
 \\
& \textbf{No SI} & \textbf{SI} & \textbf{No SI} & \textbf{SI}
 \\
\hline

\multicolumn{5}{|c|}{Zeroshot} 
\\ 
\hline
 mTOP(de) & 60.27 $\pm$ 1.40 & \textbf{65.15} $\pm$ 0.88 & 60.66 $\pm$ 1.77 & 62.89 $\pm$ 1.63 
 \\

 mTOP(th) & 43.30 $\pm$ 0.75 & 43.45 $\pm$ 1.16 & 44.41 $\pm$ 1.45 & \textbf{44.61} $\pm$ 2.58
 \\

 \hline 

\multicolumn{5}{|c|}{Joint Multilingual} 
\\
\hline

 mTOP(de) & \textbf{79.47} $\pm$ 0.45 & 78.77 $\pm$ 0.47 & 78.44 $\pm$ 0.66 & 77.77 $\pm$ 0.48
\\ 
 mTOP(th) & \textbf{77.58} $\pm$ 0.14 & 77.32 $\pm$ 0.84 & 76.19 $\pm$ 0.61 & 75.36 $\pm$ 0.64
 \\
 \hline

 \end{tabular}
 }

 \caption{Scores for the \textbf{Perfect} metric with and without spaces added pre-tokenization of the input of the \textbf{Sentinel + Tag} approaches.}
   \label{tab:multilingual-space-sentinel}
 \end{table}

\section{Robustness of decoder distribution}


While the results so far have focused on the top-1 prediction, in many applications we want a robust and meaningful top-$K$ prediction set.
Thus we investigated the robustness of the decoder's output distribution.
Table~\ref{tab:topk_analysis} shows the \% of the test examples whose top-$K$ predictions (produced via beam search with $K=5$) include the ground-truth label.

\begin{table}[!h]
\centering
 \resizebox{\textwidth / 2}{!}{
 \begin{tabular}{|l|c|c|c|c|c|}
 \toprule
\multirow{1}{*}{\textbf{Method}} & \multicolumn{1}{c|}{\textbf{mATIS}(en)} & \multicolumn{1}{c|}{\textbf{SNIPS}} & \multicolumn{1}{c|}{\textbf{MovieTrivia}} & \multicolumn{1}{c|}{\textbf{Movies}} & \multicolumn{1}{c|}{\textbf{Restaurant}}
\\
  \toprule

Tag Only & 92.16 & 96.71 & 47.47 & 88.82 & 85.03  
 \\ \hline
Input + Tag & 90.03 & 96.29 & 48.85 & 90.45 & 86.61 
 \\ \hline
Tagged Spans & 88.24 & 96.86 & 69.53 & 88.24 & 87.40 
 \\ \hline
Sentinel + Tag & 93.84 & 98.29 & 67.95 & \textbf{90.82} & 86.28 
 \\ \hline
\hline 
\multicolumn{6}{|c|}{\textbf{Simplified Inside Based Formats}}
 \\ \hline 
Tag Only (SI) & 92.39 & 96.57 & 62.98 & 89.88 & 84.23 
 \\ \hline
Input + Tag (SI) & 91.94 & 97.14 & 66.72 & 90.66 & 87.34 
 \\ \hline
Sentinel + Tag (SI) & \textbf{94.40} & \textbf{98.71} & \textbf{72.20} & 90.78 & \textbf{87.53} 
 \\ 
 \bottomrule
 \end{tabular}
}

 \caption{\% of examples with any of the top-$5$ predictions containing the ground-truth label.}
   \label{tab:topk_analysis}
 \end{table}

As expected, the observed scores are significantly higher than the perfect accuracy scores (for the Top-1) seen in Table~\ref{tab:base-128-256-test}.
Also, consistent with previous results, the \textbf{Sentinel + Tag (SI)} performed the best on average.
In general, this analysis encourages us to think about how we could leverage these instances (where the correct prediction is in the top-$K$ but not the top-1) to further improve performance. We leave this for future work.


\begin{table*}[!t]
\centering
 \resizebox{\textwidth}{!}{
 \begin{tabular}{|l|llllllll|}
 \hline
 
\textbf{Format} & \multicolumn{8}{l|}{\textbf{Sample \textit{Target} }} \\
  \hline

\multicolumn{9}{|c|}{\textbf{Simplified Inside Based Formats}} \\
\hline

\textbf{Tag Only (SI)} & 
$\targetsi$: & O &  ARTIST & I & O & O & PLAYLIST & O \\
\hline

\textbf{Tag + Input (SI)} & 
$\targetsi$: & <O> Add & <ARTIST> Kent & <I> James & <O> to & <O> the & <PLAYLIST> Disney & <O> soundtrack \\
\hline

\textbf{Sentinel + Tag (SI)} & 
$\targetsi$: &  <extra\_id\_0> O & <extra\_id\_1> ARTIST  & <extra\_id\_2> I & <extra\_id\_3> O & <extra\_id\_4> O & <extra\_id\_5> PLAYLIST & <extra\_id\_6> O \\
\hline

\multicolumn{9}{|c|}{\textbf{Simplified Outside Based Formats}} \\
\hline

\textbf{Tag + Input (SI, SO)} & 
$\targetsi$: & Add & <ARTIST> Kent & <I> James & to  & the & <PLAYLIST> Disney  & soundtrack \\
\hline

\textbf{Tagged Spans (SO)} & 
$\targetsi$: & Add & <ARTIST> Kent & James </> & to & the & <PLAYLIST> Disney </> & soundtrack \\
\hline

\textbf{Sentinel + Tag (SI, SO)} & 
$\targetsi$: & <extra\_id\_0> & <extra\_id\_1> ARTIST & <extra\_id\_2> I & <extra\_id\_3> & <extra\_id\_4> & <extra\_id\_5> PLAYLIST & <extra\_id\_6> \\
\hline
\hline
\multicolumn{9}{|c|}{\textbf{Extractive Sentinel + Tag (ES+T) Based Formats}} \\
\hline

\textbf{ES+T} & 
$\targetsi$: & & <extra\_id\_1> ARTIST & <extra\_id\_2> I-ARTIST & & & <extra\_id\_5> PLAYLIST & \\
\hline

\textbf{ES+T (Simplified)} & 
$\targetsi$: & & <extra\_id\_1> ARTIST & <extra\_id\_2> & & & <extra\_id\_5> PLAYLIST & \\
\hline

\end{tabular}
 }

 \caption{Examples for other format variants we studied. For all the non-sentinel approaches, examples, the $\inputsi$ is the original input utterance "Add Kent James to the Disney soundtrack". For sentinel-based approaches, the $\inputsi$ is "<extra\_id\_0> Add <extra\_id\_1> Kent <extra\_id\_2> James <extra\_id\_3> to <extra\_id\_4> the <extra\_id\_5> Disney <extra\_id\_6> soundtrack". }
 \label{tab:format-variant-examples}
 \end{table*}

\begin{table*}[!ht]
\centering
 \resizebox{\textwidth}{!}{
 \begin{tabular}{|l|cccc||cccc||cccc||cccc|}
\hline
\multirow{2}{*}{Size} & \multicolumn{4}{c||}{\textbf{Average (all) [ZS]}} & \multicolumn{4}{c||}{\textbf{Average (non-en) [ZS]}} & \multicolumn{4}{c|}{\textbf{Average (all) [J]}} & \multicolumn{4}{c|}{\textbf{Average (non-en) [J]}}
 \\
& \textbf{T-O} & \textbf{I+T} & \textbf{TS} & \textbf{S+T} & \textbf{T-O} & \textbf{I+T} & \textbf{TS} & \textbf{S+T} & \textbf{T-O} & \textbf{I+T} & \textbf{TS} & \textbf{S+T} & \textbf{T-O} & \textbf{I+T} & \textbf{TS} & \textbf{S+T}
 \\
\hline 

In: 128 Out: 256
& 48.68 & 49.54 & 45.58 & \textbf{57.79}
& 35.51 & 36.47 & 31.29 & \textbf{47.11}
& 76.63 & 78.11 & 76.15 & \textbf{81.21}
& 73.22 & 74.89 & 72.60 & \textbf{78.78}
 \\ 
\hline

In: 128 Out: 128
& 48.79 & 49.55 & 46.15 & \textbf{55.35}
& 35.59 & 36.65 & 32.26 & \textbf{44.21}
& 75.95 & 75.08 & 75.22 & \textbf{79.89}
& 72.51 & 71.83 & 71.69 & \textbf{77.22}
 \\ 
\hline

In: 64 Out: 64
& 47.01 & 48.52 & 45.23 & \textbf{54.89}
& 33.84 & 35.91 & 32.06 & \textbf{43.82}
& 72.91 & 74.92 & 71.11 & \textbf{79.95}
& 69.45 & 71.96 & 67.66 & \textbf{77.61}
 \\ 
\hline

 \hline
 \end{tabular}
 }

  \caption{Results demonstrating impact of changing training (input and output / target) sequence lengths for Base models in a similar setup as Table~\ref{tab:multilingual-model-size}.}
   \label{tab:multilingual-seq-len}
 \end{table*}


\begin{table*}[!h]
\centering
 \resizebox{\textwidth}{!}{
\begin{tabular}{|l|c|c|c|c|c|c|c|c|c|c|c|c|c|c|c|c|}
 \hline
 
\multirow{2}{*}{\textbf{Format}} & \multicolumn{3}{|c|}{\textbf{mATIS}} & \textbf{SNIPS} & \textbf{MovieTrivia} & \textbf{Movies} & \textbf{Restaurant} & \multicolumn{6}{|c|}{\textbf{mTOP}} &  \multicolumn{3}{|c|}{\textbf{mTOD}}
\\

& en & hi & tr & & & & & en & de & es & fr & hi & th & en & es & th
\\
  \hline
  
  \multicolumn{17}{|c|}{\textbf{Input Formats}} \\ \hline
  
Original Input & 16 / 37 & 24 / 54 & 20 / 42 & 12 / 26 & 28 / 56 & 13 / 32 & 12 / 27 & 9 / 23 & 11 / 28 & 12 / 31 & 13 / 33 & 15 / 37 & 10 / 24 & 9 / 20 & 11 / 24 & 13 / 28 \\ \hline
Sentinel Input & 28 / 61 & 36 / 83 & 31 / 64 & 21 / 44 & 49 / 95 & 24 / 55 & 21 / 48 & 17 / 43 & 20 / 51 & 23 / 56 & 23 / 56 & 24 / 58 & 24 / 60 & 16 / 34 & 19 / 40 & 21 / 44 \\ 
Sentinel Input (NoSpace) &  &  &  &  &  &  &  & 17 / 41 & 19 / 49 & 22 / 54 & 23 / 54 & 24 / 57 & 20 / 49 &  &  &  \\ \hline

\multicolumn{17}{|c|}{\textbf{Target Formats}} \\ \hline

Tag Only & 31 / 76 & 30 / 76 & 28 / 68 & 20 / 54 & 50 / 120 & 18 / 49 & 14 / 38 & 19 / 74 & 19 / 79 & 22 / 85 & 21 / 83 & 19 / 72 & 25 / 91 & 14 / 47 & 17 / 55 & 18 / 59 \\ 
Tag Only (SI) & 26 / 59 & 26 / 59 & 24 / 53 & 13 / 28 & 21 / 42 & 13 / 32 & 10 / 22 & 12 / 33 & 13 / 33 & 13 / 32 & 13 / 33 & 13 / 33 & 13 / 32 & 10 / 22 & 10 / 21 & 10 / 21 \\ \hline
Input + Tag & 71 / 157 & 78 / 177 & 70 / 156 & 50 / 111 & 120 / 249 & 52 / 124 & 46 / 103 & 43 / 129 & 47 / 144 & 53 / 159 & 53 / 154 & 51 / 144 & 58 / 167 & 37 / 89 & 43 / 104 & 47 / 108 \\ 
Input + Tag (SI) & 66 / 144 & 74 / 161 & 67 / 148 & 43 / 87 & 91 / 176 & 47 / 108 & 41 / 92 & 37 / 86 & 40 / 98 & 45 / 106 & 45 / 104 & 45 / 105 & 47 / 111 & 33 / 67 & 36 / 75 & 39 / 79 \\ 
Input + Tag (SI, SO) & 45 / 101 & 51 / 116 & 45 / 103 & 29 / 64 & 69 / 145 & 28 / 67 & 24 / 55 & 24 / 70 & 26 / 80 & 29 / 87 & 29 / 83 & 30 / 83 & 32 / 91 & 21 / 50 & 25 / 61 & 26 / 61 \\ \hline
Tagged Spans & 85 / 187 & 93 / 202 & 85 / 194 & 50 / 108 & 80 / 167 & 58 / 138 & 52 / 117 & 44 / 90 & 47 / 100 & 51 / 102 & 54 / 111 & 53 / 109 & 51 / 103 & 40 / 81 & 39 / 80 & 43 / 86 \\ 
Tagged Spans (SO) & 49 / 109 & 55 / 120 & 49 / 111 & 29 / 59 & 44 / 82 & 27 / 65 & 24 / 47 & 22 / 54 & 25 / 60 & 26 / 58 & 26 / 60 & 28 / 65 & 27 / 61 & 20 / 49 & 21 / 43 & 22 / 44 \\ \hline
Sentinel + Tag & 42 / 98 & 41 / 99 & 38 / 90 & 29 / 70 & 71 / 156 & 29 / 71 & 23 / 57 & 26 / 91 & 27 / 98 & 31 / 105 & 30 / 103 & 27 / 90 & 34 / 113 & 21 / 60 & 24 / 69 & 26 / 72 \\ 
Sentinel + Tag (SI) & 38 / 83 & 37 / 82 & 35 / 74 & 22 / 45 & 42 / 82 & 23 / 55 & 19 / 43 & 20 / 47 & 21 / 50 & 22 / 52 & 22 / 52 & 21 / 50 & 23 / 53 & 17 / 35 & 17 / 37 & 18 / 37 \\ 
Sentinel + Tag (SI, SO) & 30 / 68 & 29 / 70 & 28 / 63 & 18 / 37 & 34 / 69 & 17 / 41 & 13 / 30 & 16 / 42 & 16 / 43 & 17 / 45 & 17 / 44 & 16 / 42 & 18 / 46 & 13 / 30 & 14 / 32 & 14 / 31 \\ \hline
Extractive Sentinel + Tag & 28 / 74 & 26 / 75 & 24 / 67 & 20 / 59 & 56 / 145 & 16 / 48 & 12 / 40 & 18 / 83 & 18 / 84 & 21 / 93 & 19 / 91 & 17 / 79 & 25 / 102 & 13 / 52 & 17 / 64 & 17 / 65 \\ 
Extractive Sentinel + Tag (S) & 22 / 55 & 21 / 52 & 20 / 51 & 11 / 26 & 17 / 36 & 9 / 24 & 6 / 15 & 10 / 31 & 10 / 32 & 10 / 29 & 9 / 28 & 9 / 31 & 10 / 30 & 7 / 23 & 8 / 21 & 7 / 20 \\ \hline
\end{tabular}
 }

 \caption{Detailed per-dataset / language statistics for the Mean and $99\%^{th}$ percentile of sequence length (\emph{i.e.,} the number SentencePiece tokens) for the different $\inputsi$ and $\targetsi$ formats studied.}
 \label{tab:seq-len-stats-full}
 \end{table*}

 \begin{table}[!h]
\centering
 \resizebox{\textwidth / 2}{!}{
 \begin{tabular}{|l|c|cc|cc|cc|}
\hline
\multirow{2}{*}{Dataset} & \multirow{2}{*}{Language} &  \multicolumn{2}{c|}{\textbf{Extractive TS}} & \multicolumn{2}{c|}{\textbf{Extractive S+T}} & \multicolumn{2}{c|}{\textbf{Extractive S+T(S)}}
 \\
& & \textbf{Perfect} & \textbf{Mic F1} & \textbf{Perfect} & \textbf{Mic F1} & \textbf{Perfect} & \textbf{Mic F1} 
 \\
\hline 

\multirow{3}{*}{\textbf{mATIS}} & en & \textbf{89.59} & \textbf{95.60} & 88.13 & 95.39 & 89.03 & 95.59
\\ 
  &  & $\pm$ 0.11 & $\pm$ 0.03 & $\pm$ 1.12 & $\pm$ 0.35 & $\pm$ 0.11 & $\pm$ 0.01
 \\
 & hi & 7.39 & 30.33 & 22.28 & 57.60 & \textbf{32.87} & \textbf{63.70}
\\ 
  &  & $\pm$ 0.90 & $\pm$ 2.70 & $\pm$ 6.72 & $\pm$ 3.06 & $\pm$ 2.30 & $\pm$ 1.50
 \\ 
 & tr & \textbf{8.53} & \textbf{46.74} & 2.45 & 39.23 & 4.62 & 43.16
\\ 
  &  & $\pm$ 0.14 & $\pm$ 0.20 & $\pm$ 0.77 & $\pm$ 2.99 & $\pm$ 0.56 & $\pm$ 0.47
 \\ \hline
\multirow{3}{*}{\textbf{mTOD}} & en & \textbf{92.23} & \textbf{96.07} & 91.62 & 95.72 & 91.45 & 95.64
\\ 
  &  & $\pm$ 0.34 & $\pm$ 0.16 & $\pm$ 0.59 & $\pm$ 0.32 & $\pm$ 0.06 & $\pm$ 0.01
 \\ 
 & es & 45.20 & 60.74 & 59.00 & 72.16 & \textbf{62.72} & \textbf{75.57}
\\ 
  &  & $\pm$ 5.64 & $\pm$ 4.97 & $\pm$ 1.00 & $\pm$ 2.16 & $\pm$ 1.50 & $\pm$ 0.99
 \\ 
 & th & 21.10 & 42.58 & 29.70 & \textbf{50.36} & \textbf{30.23} & 49.76
\\ 
  &  & $\pm$ 1.83 & $\pm$ 2.29 & $\pm$ 4.76 & $\pm$ 1.15 & $\pm$ 0.27 & $\pm$ 0.20
 \\ \hline
\multirow{6}{*}{\textbf{mTOP}} & en & 87.22 & 88.60 & 87.24 & 87.20 & \textbf{88.36} & \textbf{88.88}
\\ 
  &  & $\pm$ 0.19 & $\pm$ 0.17 & $\pm$ 1.63 & $\pm$ 1.81 & $\pm$ 0.72 & $\pm$ 0.47
 \\ 
 & de & 73.68 & 59.91 & 80.70 & 71.28 & \textbf{81.50} & \textbf{73.29}
\\ 
  &  & $\pm$ 1.01 & $\pm$ 2.37 & $\pm$ 1.55 & $\pm$ 3.60 & $\pm$ 1.93 & $\pm$ 2.06
 \\ 
 & es & 83.79 & 55.52 & 89.39 & 68.70 & \textbf{89.44} & \textbf{69.26}
\\ 
  &  & $\pm$ 0.43 & $\pm$ 1.48 & $\pm$ 0.13 & $\pm$ 1.76 & $\pm$ 0.78 & $\pm$ 2.71
 \\ 
 & fr & 63.14 & 51.49 & 80.02 & 64.05 & \textbf{81.01} & \textbf{67.29}
\\ 
  &  & $\pm$ 1.85 & $\pm$ 3.27 & $\pm$ 1.47 & $\pm$ 3.15 & $\pm$ 1.39 & $\pm$ 2.08
 \\ 
 & hi & 44.25 & 33.24 & 43.64 & 32.02 & \textbf{44.69} & \textbf{34.16}
\\ 
  &  & $\pm$ 1.47 & $\pm$ 3.25 & $\pm$ 1.22 & $\pm$ 1.64 & $\pm$ 1.27 & $\pm$ 2.13
 \\ 
 & th & 40.52 & 1.58 & \textbf{99.86} & \textbf{80.95} & 99.84 & 78.57
\\ 
  &  & $\pm$ 1.97 & $\pm$ 0.19 & $\pm$ 0.00 & $\pm$ 0.00 & $\pm$ 0.02 & $\pm$ 2.38
 \\ \hline
\multicolumn{2}{|c|}{\textbf{Average}} & 54.72 & 55.20 & 64.50 & 67.89 & \textbf{66.31} & \textbf{69.57}
\\ 
\multicolumn{2}{|c|}{}  & $\pm$ 0.17 & $\pm$ 0.06 & $\pm$ 1.48 & $\pm$ 1.42 & $\pm$ 0.83 & $\pm$ 0.82
 \\ \hline
 \multicolumn{2}{|c|}{\textbf{Average}} & 43.07 & 42.46 & 56.34 & 59.60 & \textbf{58.55} & \textbf{61.64}
\\ 
 \multicolumn{2}{|c|}{ \textbf{(non-en)}}  & $\pm$ 0.20 & $\pm$ 0.09 & $\pm$ 1.74 & $\pm$ 1.69 & $\pm$ 1.05 & $\pm$ 1.04
 \\ \hline

 \end{tabular}
 }

 \caption{Detailed per-language \textbf{extractive} results for \textbf{Zero-shot} setting (mean and stdev over 2 runs).}
   \label{tab:multilingual-results-zs-extractive-full}
 \end{table}

 \begin{table}[!h]
\centering
 \resizebox{\textwidth / 2}{!}{
 \begin{tabular}{|l|c|cc|cc|cc|}
\hline
\multirow{2}{*}{Dataset} & \multirow{2}{*}{Language} &  \multicolumn{2}{c|}{\textbf{Extractive TS}} & \multicolumn{2}{c|}{\textbf{Extractive S+T}} & \multicolumn{2}{c|}{\textbf{Extractive S+T(S)}}
 \\
& & \textbf{Perfect} & \textbf{Mic F1} & \textbf{Perfect} & \textbf{Mic F1} & \textbf{Perfect} & \textbf{Mic F1} 
 \\
\hline 

\multirow{3}{*}{\textbf{mATIS}} & en & \textbf{89.42} & \textbf{95.71} & 87.23 & 94.98 & 89.25 & 95.65
\\ 
  &  & $\pm$ 0.06 & $\pm$ 0.10 & $\pm$ 1.57 & $\pm$ 0.43 & $\pm$ 0.34 & $\pm$ 0.12
 \\
 & hi & 58.17 & 81.21 & 64.95 & 84.39 & \textbf{65.73} & \textbf{84.62}
\\ 
  &  & $\pm$ 2.74 & $\pm$ 1.38 & $\pm$ 0.90 & $\pm$ 0.45 & $\pm$ 1.46 & $\pm$ 1.00
 \\ 
 & tr & 62.45 & 83.07 & \textbf{63.99} & 83.88 & 63.85 & \textbf{84.00}
\\ 
  &  & $\pm$ 0.91 & $\pm$ 0.16 & $\pm$ 0.77 & $\pm$ 0.76 & $\pm$ 0.77 & $\pm$ 0.21
 \\ \hline
\multirow{3}{*}{\textbf{mTOD}} & en & \textbf{91.46} & \textbf{95.69} & 90.08 & 94.85 & 91.06 & 95.31
\\ 
  &  & $\pm$ 0.15 & $\pm$ 0.09 & $\pm$ 0.64 & $\pm$ 0.38 & $\pm$ 0.15 & $\pm$ 0.08
 \\ 
 & es & 84.60 & 89.89 & 83.96 & 89.58 & \textbf{84.97} & \textbf{90.04}
\\ 
  &  & $\pm$ 0.18 & $\pm$ 0.07 & $\pm$ 0.43 & $\pm$ 0.31 & $\pm$ 0.35 & $\pm$ 0.15
 \\ 
 & th & \textbf{90.99} & \textbf{94.36} & 88.39 & 92.66 & 89.72 & 93.31
\\ 
  &  & $\pm$ 0.15 & $\pm$ 0.05 & $\pm$ 1.03 & $\pm$ 0.50 & $\pm$ 0.41 & $\pm$ 0.49
 \\ \hline
\multirow{6}{*}{\textbf{mTOP}} & en & 82.66 & 85.07 & 87.76 & 88.24 & \textbf{90.24} & \textbf{90.56}
\\ 
  &  & $\pm$ 0.70 & $\pm$ 0.69 & $\pm$ 1.64 & $\pm$ 1.40 & $\pm$ 0.05 & $\pm$ 0.09
 \\ 
 & de & 87.31 & 83.61 & 90.73 & 86.71 & \textbf{92.05} & \textbf{88.85}
\\ 
  & & $\pm$ 0.38 & $\pm$ 0.66 & $\pm$ 0.87 & $\pm$ 1.20 & $\pm$ 0.08 & $\pm$ 0.08
 \\ 
 & es & 92.43 & 81.57 & 94.31 & 84.41 & \textbf{95.15} & \textbf{86.47}
\\ 
  &  & $\pm$ 0.00 & $\pm$ 0.31 & $\pm$ 0.72 & $\pm$ 1.73 & $\pm$ 0.15 & $\pm$ 0.40
 \\ 
 & fr & 87.97 & 82.09 & 89.59 & 83.05 & \textbf{90.82} & \textbf{84.89}
\\ 
  &  & $\pm$ 0.25 & $\pm$ 0.42 & $\pm$ 0.92 & $\pm$ 1.18 & $\pm$ 0.19 & $\pm$ 0.47
 \\ 
 & hi & 77.02 & 79.40 & 81.19 & 82.42 & \textbf{83.24} & \textbf{84.42}
\\ 
  &  & $\pm$ 0.47 & $\pm$ 0.55 & $\pm$ 1.13 & $\pm$ 1.16 & $\pm$ 0.13 & $\pm$ 0.00
 \\ 
 & th & 99.29 & 51.29 & \textbf{99.86} & 80.07 & \textbf{99.86} & \textbf{80.95}
\\ 
  &  & $\pm$ 0.02 & $\pm$ 0.44 & $\pm$ 0.04 & $\pm$ 5.65 & $\pm$ 0.04 & $\pm$ 4.76
 \\ \hline
\multicolumn{2}{|c|}{\textbf{Average}} & 83.65 & 83.58 & 85.17 & 87.10 & \textbf{86.33} & \textbf{88.26}
\\ 
\multicolumn{2}{|c|}{}  & $\pm$ 0.49 & $\pm$ 0.38 & $\pm$ 0.53 & $\pm$ 0.12 & $\pm$ 0.16 & $\pm$ 0.54
 \\ \hline
 \multicolumn{2}{|c|}{\textbf{Average}} & 82.25 & 80.72 & 84.11 & 85.24 & \textbf{85.04} & \textbf{86.40}
\\ 
 \multicolumn{2}{|c|}{ \textbf{(non-en)}}  & $\pm$ 0.57 & $\pm$ 0.43 & $\pm$ 0.42 & $\pm$ 0.00 & $\pm$ 0.16 & $\pm$ 0.71
 \\ \hline

 \end{tabular}
 }

 \caption{Per-language \textbf{extractive} results for \textbf{joint multilingual} setting (mean and stdev over 2 runs).}
   \label{tab:multilingual-results-joint-extractive-full}
 \end{table}

 \begin{table}[!h]
\centering
 \resizebox{\textwidth / 2}{!}{
 \begin{tabular}{l|c|ccc||ccc}
\hline
\multirow{2}{*}{Dataset} & \multirow{2}{*}{Lang} & \multicolumn{3}{c||}{\textbf{Zeroshot}} & \multicolumn{3}{c}{\textbf{Joint}}
 \\
& & \textbf{E TS} & \textbf{E S+T} & \textbf{E S+T(S)} & \textbf{E TS} & \textbf{E S+T} & \textbf{E S+T(S)}
 \\
\hline 

\multirow{3}{*}{\textbf{mATIS}} & en & 0.56 & 0.11 & 0.00 & 0.34 & 0.00 & 0.11
\\ 
  &  & $\pm$ 0.11 & $\pm$ 0.11 & $\pm$ 0.00 & $\pm$ 0.00 & $\pm$ 0.00 & $\pm$ 0.11
 \\
 & hi & 79.84 & 0.62 & 0.17 & 9.69 & 0.06 & 0.00
\\ 
  &  & $\pm$ 3.47 & $\pm$ 0.39 & $\pm$ 0.06 & $\pm$ 1.06 & $\pm$ 0.06 & $\pm$ 0.00
 \\ 
 & tr & 17.55 & 0.49 & 0.35 & 2.24 & 0.14 & 0.07
\\ 
  &  & $\pm$ 1.89 & $\pm$ 0.49 & $\pm$ 0.21 & $\pm$ 0.14 & $\pm$ 0.00 & $\pm$ 0.07
 \\ \hline
\multirow{3}{*}{\textbf{mTOD}} & en & 0.04 & 0.01 & 0.01 & 0.03 & 0.11 & 0.00
\\ 
  &  & $\pm$ 0.03 & $\pm$ 0.01 & $\pm$ 0.01 & $\pm$ 0.01 & $\pm$ 0.02 & $\pm$ 0.00
 \\ 
 & es & 33.21 & 0.00 & 0.02 & 0.20 & 0.00 & 0.00
\\ 
  &  & $\pm$ 5.44 & $\pm$ 0.00 & $\pm$ 0.02 & $\pm$ 0.03 & $\pm$ 0.00 & $\pm$ 0.00
 \\ 
 & th & 40.51 & 0.35 & 0.33 & 0.71 & 0.21 & 0.21
\\ 
  &  & $\pm$ 0.38 & $\pm$ 0.24 & $\pm$ 0.15 & $\pm$ 0.12 & $\pm$ 0.03 & $\pm$ 0.03
 \\ \hline
\multirow{6}{*}{\textbf{mTOP}} & en & 0.49 & 0.27 & 0.09 & 0.57 & 0.00 & 0.00
\\ 
  &  & $\pm$ 0.15 & $\pm$ 0.23 & $\pm$ 0.02 & $\pm$ 0.36 & $\pm$ 0.00 & $\pm$ 0.00
 \\ 
 & de & 9.85 & 0.39 & 0.03 & 0.23 & 0.00 & 0.00
\\ 
  & & $\pm$ 0.97 & $\pm$ 0.25 & $\pm$ 0.03 & $\pm$ 0.08 & $\pm$ 0.00 & $\pm$ 0.00
 \\ 
 & es & 5.44 & 0.17 & 0.00 & 0.22 & 0.03 & 0.00
\\ 
  &  & $\pm$ 0.40 & $\pm$ 0.13 & $\pm$ 0.00 & $\pm$ 0.12 & $\pm$ 0.03 & $\pm$ 0.00
 \\ 
 & fr & 9.10 & 0.25 & 0.03 & 0.17 & 0.02 & 0.00
\\ 
  &  & $\pm$ 1.46 & $\pm$ 0.22 & $\pm$ 0.00 & $\pm$ 0.08 & $\pm$ 0.02 & $\pm$ 0.00
 \\ 
 & hi & 14.34 & 0.02 & 0.00 & 2.17 & 0.00 & 0.00 
\\ 
  &  & $\pm$ 1.43 & $\pm$ 0.02 & $\pm$ 0.00 & $\pm$ 0.05 & $\pm$ 0.00 & $\pm$ 0.00
 \\ 
 & th & 66.89 & 0.00 & 0.00 & 0.02 & 0.00 & 0.00
\\ 
  &  & $\pm$ 0.60 & $\pm$ 0.00 & $\pm$ 0.00 & $\pm$ 0.02 & $\pm$ 0.00 & $\pm$ 0.00
 \\ \hline
\multicolumn{2}{c|}{\textbf{Average}} & 23.15 & 0.22 & 0.09 & 1.38 & 0.05 & 0.03
\\ 
\multicolumn{2}{c|}{}  & $\pm$ 0.61 & $\pm$ 0.16 & $\pm$ 0.01 & $\pm$ 0.16 & $\pm$ 0.01 & $\pm$ 0.02
 \\ \hline
 \multicolumn{2}{c|}{\textbf{Average}} & 30.75 & 0.25 & 0.10 & 1.74 & 0.05 & 0.03
\\ 
 \multicolumn{2}{c|}{ \textbf{(non-en)}}  & $\pm$ 0.84 & $\pm$ 0.19 & $\pm$ 0.01 & $\pm$ 0.18 & $\pm$ 0.01 & $\pm$ 0.01
 \\ \hline

 \end{tabular}
 }

 \caption{\% of hallucination per dataset (mean and stdev over 2 runs) for \textbf{extractive} formats.}
   \label{tab:multilingual-hallucinations-extractive-full}
 \end{table}


 \begin{table*}[h]
\centering
 \resizebox{\textwidth}{!}{
 \begin{tabular}{|l|c|cc|cc||cc|cc||cc||cc|cc|}
\hline
\multirow{2}{*}{Dataset} & \multirow{2}{*}{Language} & \multicolumn{2}{c|}{\textbf{Tag Only}} & \multicolumn{2}{c||}{\textbf{Tag Only (SI)}} & \multicolumn{2}{c|}{\textbf{Input + Tag}} & \multicolumn{2}{c||}{\textbf{Input + Tag (SI)}} & \multicolumn{2}{c||}{\textbf{Tagged Spans}} & \multicolumn{2}{c|}{\textbf{Sentinel + Tag}} & \multicolumn{2}{c|}{\textbf{Sentinel + Tag (SI)}}
 \\
& & \textbf{Perfect} & \textbf{Mic F1} & \textbf{Perfect} & \textbf{Mic F1} & \textbf{Perfect} & \textbf{Mic F1} & \textbf{Perfect} & \textbf{Mic F1} & \textbf{Perfect} & \textbf{Mic F1} & \textbf{Perfect} & \textbf{Mic F1} & \textbf{Perfect} & \textbf{Mic F1}
 \\
\hline 

\multirow{3}{*}{\textbf{mATIS}} & en & 87.83 & 93.84 & 88.39 & 94.27 & 88.84 & 95.40 & 88.69 & 95.36 & 88.99 & 95.54 & 89.32 & 95.71 & \textbf{89.77} & \textbf{95.87}
\\ 
  &  & $\pm$ 1.42 & $\pm$ 0.90 & $\pm$ 0.67 & $\pm$ 0.74 & $\pm$ 0.28 & $\pm$ 0.17 & $\pm$ 0.33 & $\pm$ 0.16 & $\pm$ 0.71 & $\pm$ 0.19 & $\pm$ 0.11 & $\pm$ 0.08 & $\pm$ 0.45 & $\pm$ 0.18
 \\ 

 & hi & 18.07 & 32.70 & 18.03 & 33.19 & 22.58 & 50.30 & 24.30 & 52.03 & 6.79 & 34.38 & \textbf{34.68} & \textbf{66.33} & 33.26 & 64.85
\\ 
  &  & $\pm$ 7.22 & $\pm$ 9.43 & $\pm$ 1.74 & $\pm$ 3.26 & $\pm$ 3.59 & $\pm$ 5.50 & $\pm$ 3.61 & $\pm$ 3.83 & $\pm$ 2.61 & $\pm$ 5.87 & $\pm$ 3.85 & $\pm$ 1.96 & $\pm$ 5.69 & $\pm$ 5.04
 \\ 

 & tr & 4.24 & 31.85 & 4.48 & 32.44 & 5.45 & 42.15 & 5.41 & 42.86 & \textbf{10.96} & \textbf{45.73} & 6.85 & 43.58 & 4.43 & 39.30
\\ 
  &  & $\pm$ 1.45 & $\pm$ 2.70 & $\pm$ 1.05 & $\pm$ 1.86 & $\pm$ 0.64 & $\pm$ 0.80 & $\pm$ 0.40 & $\pm$ 0.46 & $\pm$ 1.83 & $\pm$ 2.62 & $\pm$ 2.36 & $\pm$ 3.82 & $\pm$ 0.54 & $\pm$ 1.92
 \\ 

 \hline 

\multirow{6}{*}{\textbf{mTOP}} & en & 81.98 & 88.50 & 83.06 & 89.59 & 84.34 & 90.80 & 84.88 & 91.20 & 84.22 & 90.82 & 85.58 & 91.68 & \textbf{86.56} & \textbf{92.28}
\\ 
  &  & $\pm$ 0.34 & $\pm$ 0.34 & $\pm$ 0.48 & $\pm$ 0.31 & $\pm$ 0.45 & $\pm$ 0.39 & $\pm$ 0.21 & $\pm$ 0.06 & $\pm$ 0.66 & $\pm$ 0.46 & $\pm$ 0.58 & $\pm$ 0.42 & $\pm$ 0.69 & $\pm$ 0.44
 \\ 

 & de & 51.56 & 65.79 & 52.99 & 66.80 & 57.16 & 71.24 & 59.36 & 73.10 & 50.00 & 64.54 & 60.27 & 75.10 & \textbf{65.15} & \textbf{77.40}
\\ 
  &  & $\pm$ 0.86 & $\pm$ 0.87 & $\pm$ 0.75 & $\pm$ 0.54 & $\pm$ 1.49 & $\pm$ 1.35 & $\pm$ 0.21 & $\pm$ 0.09 & $\pm$ 1.73 & $\pm$ 1.46 & $\pm$ 1.40 & $\pm$ 0.70 & $\pm$ 0.88 & $\pm$ 0.77
 \\ 

 & es & 45.55 & 51.48 & 48.52 & 54.72 & 42.46 & 50.86 & 42.14 & 50.38 & 51.09 & 61.93 & 62.74 & 73.90 & \textbf{63.30} & \textbf{74.30}
\\ 
  &  & $\pm$ 1.50 & $\pm$ 1.49 & $\pm$ 0.82 & $\pm$ 0.81 & $\pm$ 2.82 & $\pm$ 2.36 & $\pm$ 0.78 & $\pm$ 0.20 & $\pm$ 0.63 & $\pm$ 0.48 & $\pm$ 0.97 & $\pm$ 1.47 & $\pm$ 1.27 & $\pm$ 1.64
 \\ 

 & fr & 44.43 & 50.32 & 46.96 & 53.81 & 53.00 & 61.95 & 54.49 & 64.27 & 52.01 & 63.95 & 60.51 & 72.68 & \textbf{62.88} & \textbf{74.47}
\\ 
  &  & $\pm$ 2.10 & $\pm$ 2.17 & $\pm$ 2.21 & $\pm$ 2.73 & $\pm$ 0.99 & $\pm$ 0.90 & $\pm$ 0.57 & $\pm$ 0.28 & $\pm$ 0.90 & $\pm$ 0.91 & $\pm$ 1.96 & $\pm$ 1.80 & $\pm$ 0.81 & $\pm$ 1.06
 \\ 

 & hi & 22.51 & 33.95 & 23.15 & 34.40 & 27.30 & 40.00 & 29.15 & 42.54 & 24.87 & 38.96 & 30.42 & 45.73 & \textbf{34.52} & \textbf{49.98}
\\ 
  &  & $\pm$ 1.22 & $\pm$ 1.28 & $\pm$ 0.65 & $\pm$ 0.56 & $\pm$ 1.16 & $\pm$ 1.54 & $\pm$ 1.33 & $\pm$ 1.48 & $\pm$ 1.80 & $\pm$ 1.86 & $\pm$ 1.07 & $\pm$ 2.07 & $\pm$ 2.72 & $\pm$ 4.47
 \\ 

 & th & 14.44 & 18.81 & 15.20 & 18.88 & 9.36 & 7.26 & 9.40 & 7.36 & 1.81 & 6.95 & 43.30 & 58.82 & \textbf{43.45} & \textbf{58.87}
\\ 
  &  & $\pm$ 0.51 & $\pm$ 0.34 & $\pm$ 0.62 & $\pm$ 0.52 & $\pm$ 0.09 & $\pm$ 0.37 & $\pm$ 0.13 & $\pm$ 0.47 & $\pm$ 0.08 & $\pm$ 0.34 & $\pm$ 0.75 & $\pm$ 0.57 & $\pm$ 1.16 & $\pm$ 0.69
 \\ 

 \hline 
 
 \multirow{3}{*}{\textbf{mTOD}} & en & 92.53 & 95.97 & 93.09 & 96.34 & 92.39 & 96.01 & 92.66 & 96.17 & 92.13 & 95.90 & 92.69 & 96.21 & \textbf{93.19} & \textbf{96.42}
\\ 
  &  & $\pm$ 0.10 & $\pm$ 0.05 & $\pm$ 0.06 & $\pm$ 0.04 & $\pm$ 0.07 & $\pm$ 0.04 & $\pm$ 0.03 & $\pm$ 0.01 & $\pm$ 0.08 & $\pm$ 0.08 & $\pm$ 0.13 & $\pm$ 0.08 & $\pm$ 0.17 & $\pm$ 0.08
 \\ 

 & es & 63.49 & 74.72 & 65.39 & 76.20 & 62.92 & 76.60 & 61.47 & 76.23 & 51.27 & 67.04 & 63.62 & 76.85 & \textbf{70.58} & \textbf{81.34}
\\ 
  &  & $\pm$ 4.60 & $\pm$ 4.95 & $\pm$ 3.87 & $\pm$ 2.84 & $\pm$ 2.27 & $\pm$ 1.23 & $\pm$ 0.84 & $\pm$ 0.15 & $\pm$ 4.03 & $\pm$ 2.97 & $\pm$ 3.58 & $\pm$ 2.04 & $\pm$ 3.73 & $\pm$ 1.60
 \\ 

 & th & 43.22 & 52.12 & 44.86 & 53.72 & 40.01 & 48.64 & 42.51 & 51.81 & 32.80 & 43.20 & 44.44 & 56.20 & \textbf{46.43} & \textbf{57.77}
\\ 
  &  & $\pm$ 1.19 & $\pm$ 0.77 & $\pm$ 0.68 & $\pm$ 0.85 & $\pm$ 1.27 & $\pm$ 1.80 & $\pm$ 1.32 & $\pm$ 1.12 & $\pm$ 1.57 & $\pm$ 2.08 & $\pm$ 1.02 & $\pm$ 0.75 & $\pm$ 1.98 & $\pm$ 1.92
 \\ 

 \hline 

\multicolumn{2}{|c|}{\textbf{Average}} & 47.49 & 57.50 & 48.68 & 58.70 & 48.82 & 60.93 & 49.54 & 61.94 & 45.58 & 59.08 & 56.20 & 71.07 & \textbf{57.79} & \textbf{71.90}
\\ 
 \multicolumn{2}{|c|}{} & $\pm$ 1.09 & $\pm$ 1.31 & $\pm$ 0.39 & $\pm$ 0.81 & $\pm$ 0.37 & $\pm$ 0.46 & $\pm$ 0.46 & $\pm$ 0.33 & $\pm$ 0.57 & $\pm$ 0.82 & $\pm$ 0.95 & $\pm$ 0.79 & $\pm$ 1.38 & $\pm$ 1.40
 \\ 
 \hline 
 
 \multicolumn{2}{|c|}{\textbf{Average}} & 34.17 & 45.75 & 35.51 & 47.13 & 35.58 & 49.89 & 36.47 & 51.18 & 31.29 & 47.41 & 45.20 & 63.24 & \textbf{47.11} & \textbf{64.25}
\\ 
 \multicolumn{2}{|c|}{ \textbf{(non-en)}} & $\pm$ 1.27 & $\pm$ 1.65 & $\pm$ 0.41 & $\pm$ 0.98 & $\pm$ 0.43 & $\pm$ 0.58 & $\pm$ 0.63 & $\pm$ 0.45 & $\pm$ 0.64 & $\pm$ 1.03 & $\pm$ 1.23 & $\pm$ 1.02 & $\pm$ 1.73 & $\pm$ 1.80
 \\ \hline
 \end{tabular}
 }

 \caption{Detailed per-language  results for \textbf{Zero-shot} setting (mean and standard deviation over 3 runs).}
   \label{tab:multilingual-results-zs-full}
 \end{table*}


 \begin{table*}[!h]
\centering
 \resizebox{\textwidth}{!}{
 \begin{tabular}{|l|c|cc|cc||cc|cc||cc||cc|cc|}
\hline
\multirow{2}{*}{Dataset} & \multirow{2}{*}{Language} & \multicolumn{2}{c|}{\textbf{Tag Only}} & \multicolumn{2}{c||}{\textbf{Tag Only (SI)}} & \multicolumn{2}{c|}{\textbf{Input + Tag}} & \multicolumn{2}{c||}{\textbf{Input + Tag (SI)}} & \multicolumn{2}{c||}{\textbf{Tagged Spans}} & \multicolumn{2}{c|}{\textbf{Sentinel + Tag}} & \multicolumn{2}{c|}{\textbf{Sentinel + Tag (SI)}}
 \\
& & \textbf{Perfect} & \textbf{Mic F1} & \textbf{Perfect} & \textbf{Mic F1} & \textbf{Perfect} & \textbf{Mic F1} & \textbf{Perfect} & \textbf{Mic F1} & \textbf{Perfect} & \textbf{Mic F1} & \textbf{Perfect} & \textbf{Mic F1} & \textbf{Perfect} & \textbf{Mic F1}
 \\
\hline 

\multirow{3}{*}{\textbf{mATIS}} & en & 87.61 & 93.63 & 87.94 & 93.89 & 88.99 & 95.48 & \textbf{89.36} & 95.58 & 88.80 & 95.50 & 89.29 & \textbf{95.67} & 89.10 & 95.67
\\ 
  &  & $\pm$ 0.35 & $\pm$ 0.23 & $\pm$ 0.59 & $\pm$ 0.40 & $\pm$ 0.65 & $\pm$ 0.33 & $\pm$ 0.33 & $\pm$ 0.10 & $\pm$ 0.24 & $\pm$ 0.08 & $\pm$ 0.14 & $\pm$ 0.07 & $\pm$ 0.32 & $\pm$ 0.08
 \\ 

 & hi & 59.35 & 78.17 & 62.08 & 79.55 & 61.81 & 82.56 & 61.55 & 82.53 & 59.43 & 81.93 & 64.28 & 84.29 & \textbf{67.23} & \textbf{85.26}
\\ 
  &  & $\pm$ 2.31 & $\pm$ 1.44 & $\pm$ 0.98 & $\pm$ 1.08 & $\pm$ 0.55 & $\pm$ 0.82 & $\pm$ 1.16 & $\pm$ 0.42 & $\pm$ 0.92 & $\pm$ 0.40 & $\pm$ 2.63 & $\pm$ 1.16 & $\pm$ 0.78 & $\pm$ 0.88
 \\ 

 & tr & 59.72 & 76.91 & 60.37 & 78.37 & 62.28 & 83.61 & 60.23 & 81.97 & 59.21 & 82.16 & \textbf{66.20} & \textbf{85.15} & 63.96 & 83.84
\\ 
  &  & $\pm$ 1.05 & $\pm$ 0.57 & $\pm$ 0.69 & $\pm$ 0.15 & $\pm$ 0.35 & $\pm$ 0.19 & $\pm$ 0.80 & $\pm$ 0.55 & $\pm$ 0.54 & $\pm$ 0.71 & $\pm$ 1.23 & $\pm$ 0.25 & $\pm$ 1.14 & $\pm$ 0.65
 \\ 

 \hline 

\multirow{6}{*}{\textbf{mTOP}} & en & 77.51 & 85.55 & 80.76 & 88.04 & 81.41 & 88.83 & 82.44 & 89.55 & 80.60 & 88.44 & \textbf{84.99} & \textbf{91.47} & 84.25 & 90.91
\\ 
  &  & $\pm$ 0.36 & $\pm$ 0.13 & $\pm$ 0.38 & $\pm$ 0.23 & $\pm$ 0.20 & $\pm$ 0.25 & $\pm$ 0.28 & $\pm$ 0.22 & $\pm$ 0.21 & $\pm$ 0.16 & $\pm$ 0.23 & $\pm$ 0.15 & $\pm$ 0.54 & $\pm$ 0.35
 \\ 

 & de & 72.61 & 81.72 & 75.93 & 84.61 & 75.92 & 85.15 & 76.16 & 85.48 & 73.97 & 83.71 & \textbf{79.47} & \textbf{87.77} & 78.77 & 87.39
\\ 
  &  & $\pm$ 0.77 & $\pm$ 0.65 & $\pm$ 0.07 & $\pm$ 0.02 & $\pm$ 0.19 & $\pm$ 0.20 & $\pm$ 0.23 & $\pm$ 0.06 & $\pm$ 0.91 & $\pm$ 0.73 & $\pm$ 0.45 & $\pm$ 0.27 & $\pm$ 0.47 & $\pm$ 0.40
 \\ 

 & es & 75.68 & 82.21 & 79.79 & 85.86 & 82.00 & 88.06 & 82.38 & 88.35 & 80.98 & 87.46 & \textbf{84.63} & \textbf{90.15} & 84.20 & 89.85
\\ 
  &  & $\pm$ 1.08 & $\pm$ 0.77 & $\pm$ 0.36 & $\pm$ 0.34 & $\pm$ 0.18 & $\pm$ 0.07 & $\pm$ 0.46 & $\pm$ 0.35 & $\pm$ 0.30 & $\pm$ 0.30 & $\pm$ 0.23 & $\pm$ 0.18 & $\pm$ 0.43 & $\pm$ 0.29
 \\ 

 & fr & 75.16 & 82.06 & 78.31 & 84.92 & 80.01 & 86.56 & 80.45 & 86.95 & 78.66 & 85.48 & \textbf{83.10} & \textbf{88.85} & 82.52 & 88.47
\\ 
  &  & $\pm$ 0.14 & $\pm$ 0.16 & $\pm$ 0.27 & $\pm$ 0.19 & $\pm$ 0.35 & $\pm$ 0.30 & $\pm$ 0.04 & $\pm$ 0.11 & $\pm$ 0.21 & $\pm$ 0.20 & $\pm$ 0.18 & $\pm$ 0.19 & $\pm$ 0.65 & $\pm$ 0.38
 \\ 

 & hi & 69.38 & 78.58 & 72.56 & 81.52 & 72.68 & 81.90 & 74.36 & 83.22 & 69.38 & 79.38 & \textbf{77.64} & \textbf{86.15} & 77.35 & 85.50
\\ 
  &  & $\pm$ 0.41 & $\pm$ 0.43 & $\pm$ 0.18 & $\pm$ 0.12 & $\pm$ 0.21 & $\pm$ 0.20 & $\pm$ 0.34 & $\pm$ 0.29 & $\pm$ 0.19 & $\pm$ 0.19 & $\pm$ 0.40 & $\pm$ 0.22 & $\pm$ 0.72 & $\pm$ 0.66
 \\ 

 & th & 46.24 & 53.73 & 53.09 & 61.03 & 59.24 & 68.12 & 61.51 & 69.84 & 55.15 & 68.16 & \textbf{77.58} & \textbf{85.18} & 77.32 & 84.98
\\ 
  &  & $\pm$ 1.53 & $\pm$ 1.88 & $\pm$ 0.09 & $\pm$ 0.15 & $\pm$ 1.08 & $\pm$ 0.98 & $\pm$ 1.03 & $\pm$ 0.58 & $\pm$ 0.67 & $\pm$ 0.55 & $\pm$ 0.14 & $\pm$ 0.03 & $\pm$ 0.84 & $\pm$ 0.59
 \\ 

 \hline 
 
 \multirow{3}{*}{\textbf{mTOD}} & en & 91.30 & 95.40 & 91.89 & 95.76 & 91.37 & 95.49 & 91.55 & 95.57 & 91.00 & 95.30 & \textbf{92.30} & \textbf{95.98} & 92.11 & 95.84
\\ 
  &  & $\pm$ 0.12 & $\pm$ 0.06 & $\pm$ 0.20 & $\pm$ 0.12 & $\pm$ 0.17 & $\pm$ 0.11 & $\pm$ 0.03 & $\pm$ 0.04 & $\pm$ 0.21 & $\pm$ 0.11 & $\pm$ 0.26 & $\pm$ 0.14 & $\pm$ 0.12 & $\pm$ 0.08
 \\ 

 & es & 84.63 & 89.43 & 85.55 & 90.27 & \textbf{86.19} & \textbf{90.72} & 86.07 & 90.67 & 85.29 & 90.28 & 85.77 & 90.53 & 85.77 & 90.59
\\ 
  &  & $\pm$ 0.09 & $\pm$ 0.06 & $\pm$ 0.45 & $\pm$ 0.37 & $\pm$ 0.31 & $\pm$ 0.25 & $\pm$ 0.26 & $\pm$ 0.08 & $\pm$ 0.36 & $\pm$ 0.33 & $\pm$ 0.15 & $\pm$ 0.10 & $\pm$ 0.19 & $\pm$ 0.17
 \\ 

 & th & 90.50 & 93.26 & 91.31 & 94.10 & 91.27 & 94.36 & 91.27 & 94.26 & 91.29 & 94.32 & 91.82 & \textbf{94.64} & \textbf{91.88} & 94.49
\\ 
  &  & $\pm$ 0.48 & $\pm$ 0.38 & $\pm$ 0.22 & $\pm$ 0.12 & $\pm$ 0.31 & $\pm$ 0.25 & $\pm$ 0.32 & $\pm$ 0.18 & $\pm$ 0.24 & $\pm$ 0.22 & $\pm$ 0.22 & $\pm$ 0.12 & $\pm$ 0.23 & $\pm$ 0.12
 \\ 

 \hline

\multicolumn{2}{|c|}{\textbf{Average}} & 74.14 & 82.55 & 76.63 & 84.83 & 77.76 & 86.74 & 78.11 & 87.00 & 76.15 & 86.01 & \textbf{81.42} & \textbf{89.65} & 81.21 & 89.40
\\ 
 \multicolumn{2}{|c|}{} & $\pm$ 0.18 & $\pm$ 0.13 & $\pm$ 0.08 & $\pm$ 0.05 & $\pm$ 0.09 & $\pm$ 0.15 & $\pm$ 0.25 & $\pm$ 0.14 & $\pm$ 0.12 & $\pm$ 0.16 & $\pm$ 0.38 & $\pm$ 0.14 & $\pm$ 0.10 & $\pm$ 0.10
 \\ \hline

\multicolumn{2}{|c|}{\textbf{Average}} & 70.37 & 79.56 & 73.22 & 82.25 & 74.60 & 84.56 & 74.89 & 84.81 & 72.60 & 83.65 & \textbf{78.94} & \textbf{88.08} & 78.78 & 87.82
\\ 
 \multicolumn{2}{|c|}{\textbf{(non-en)}} & $\pm$ 0.21 & $\pm$ 0.20 & $\pm$ 0.04 & $\pm$ 0.05 & $\pm$ 0.14 & $\pm$ 0.21 & $\pm$ 0.32 & $\pm$ 0.17 & $\pm$ 0.14 & $\pm$ 0.20 & $\pm$ 0.47 & $\pm$ 0.17 & $\pm$ 0.10 & $\pm$ 0.10
 \\ \hline
 \end{tabular}
 }
  \caption{Detailed per-language  results for the regular \textbf{joint multilingual} setting (mean and standard deviation over 3 runs).}
   \label{tab:multilingual-results-full}
 \end{table*}

\begin{table*}[ht]
\centering
 \resizebox{\textwidth}{!}{
 \begin{tabular}{|l|cccc||cccc||cccc||cccc|}
\hline
\multirow{2}{*}{Size} & \multicolumn{4}{c||}{\textbf{Average (all) [Zero-Shot]}} & \multicolumn{4}{c||}{\textbf{Average (non-en) [Zero-Shot]}} & \multicolumn{4}{c||}{\textbf{Average (all) [Joint]}} & \multicolumn{4}{c|}{\textbf{Average (non-en) [Joint]}}
 \\
& \textbf{T-O} & \textbf{I+T} & \textbf{TS} & \textbf{S+T} & \textbf{T-O} & \textbf{I+T} & \textbf{TS} & \textbf{S+T} & \textbf{T-O} & \textbf{I+T} & \textbf{TS} & \textbf{S+T} & \textbf{T-O} & \textbf{I+T} & \textbf{TS} & \textbf{S+T}
 \\
\hline 

Small    
& 55.60 & 57.84 & 56.74 & \textbf{62.65}\textsuperscript{\textdagger}
& 43.15 & 45.93 & 44.55 & \textbf{52.09}\textsuperscript{\textdagger}
& 83.06 & 85.04 & 84.03 & \textbf{87.53}\textsuperscript{\textdagger}
& 80.36 & 82.66 & 81.49 & \textbf{85.72}\textsuperscript{\textdagger}
\\
(300M)
& $\pm$ 0.13 & $\pm$ 0.19 & $\pm$ 0.04 & $\pm$ 0.11
& $\pm$ 0.15 & $\pm$ 0.26 & $\pm$ 0.00 & $\pm$ 0.16
& $\pm$ 0.09 & $\pm$ 0.15 & $\pm$ 0.16 & $\pm$ 0.09
& $\pm$ 0.10 & $\pm$ 0.19 & $\pm$ 0.17 & $\pm$ 0.11

\\
\hline

Base
& 58.70 & 61.94 & 59.08 & \textbf{71.90}\textsuperscript{\textdagger}
& 47.13 & 51.18 & 47.41 & \textbf{64.25}\textsuperscript{\textdagger}
& 84.83 & 87.00 & 86.01 & \textbf{89.40}\textsuperscript{\textdagger}
& 82.25 & 84.81 & 83.65 & \textbf{87.82}\textsuperscript{\textdagger}
\\ 
(580M)
& $\pm$ 0.81 & $\pm$ 0.33 & $\pm$ 0.82 & $\pm$ 1.40
& $\pm$ 0.98 & $\pm$ 0.45 & $\pm$ 1.03 & $\pm$ 1.80
& $\pm$ 0.05 & $\pm$ 0.14 & $\pm$ 0.16 & $\pm$ 0.10
& $\pm$ 0.05 & $\pm$ 0.17 & $\pm$ 0.20 & $\pm$ 0.10
 \\ 
\hline

Large
& 59.96 & 61.10 & 58.65 & \textbf{73.25}\textsuperscript{\textdagger}
& 48.94 & 49.88 & 46.67 & \textbf{66.07}\textsuperscript{\textdagger}
& 86.78 & 88.36 & 87.73 & \textbf{89.28}\textsuperscript{\textdagger}
& 84.58 & 86.42 & 85.54 & \textbf{87.62}\textsuperscript{\textdagger}
\\
(1.2B)
& $\pm$ 0.59 & $\pm$ 1.29 & $\pm$ 0.46 & $\pm$ 0.85
& $\pm$ 0.68 & $\pm$ 1.70 & $\pm$ 0.61 & $\pm$ 1.09
& $\pm$ 0.10 & $\pm$ 0.13 & $\pm$ 0.09 & $\pm$ 0.07
& $\pm$ 0.13 & $\pm$ 0.14 & $\pm$ 0.10 & $\pm$ 0.14
\\

\hline
XL
& 63.10 & 70.64 & 72.75 & \textbf{77.96}\textsuperscript{\textdagger}
& 52.50 & 62.44 & 65.23 & \textbf{72.18}\textsuperscript{\textdagger}
& 89.39 & 90.06 & 89.77 & \textbf{91.20}\textsuperscript{\textdagger}
& 87.50 & 88.31 & 87.94 & \textbf{89.79}\textsuperscript{\textdagger}

\\
(3.7B)
& $\pm$ 0.55 & $\pm$ 0.16 & $\pm$ 0.84 & $\pm$ 0.10
& $\pm$ 0.72 & $\pm$ 0.20 & $\pm$ 1.13 & $\pm$ 0.12
& $\pm$ 0.21 & $\pm$ 0.07 & $\pm$ 0.04 & $\pm$ 0.10
& $\pm$ 0.13 & $\pm$ 0.08 & $\pm$ 0.07 & $\pm$ 0.12
\\
\hline

XXL
& 61.34 & 68.66 & 73.12 & \textbf{78.27}
& 50.16 & 59.76 & 65.72 & \textbf{72.69}
& 89.06 & 90.11 & 90.13 & \textbf{91.21}
& 87.04 & 88.38 & 88.36 & \textbf{89.84}

\\
\hline

 \hline
 \end{tabular}
 }

  \caption{Impact of model size (\# of params in parentheses) on performance of formats. Averaged \textbf{MicroF1} scores are reported over the same 3 benchmarks (12 test sets) as Table~\ref{tab:multilingual-results-zs}, on both zero-shot and joint settings. The 4 methods compared are \textbf{T-O}: Tag Only (SI), \textbf{I+T}: Input + Tag (SI), \textbf{TS}: Tagged Spans and \textbf{S+T}: Sentinel + Tag (SI). Base results are averaged over 3 runs. XL, Large and Small were averaged over 2 runs. Due to compute limits, XXL (13B params) was run once for 2k steps (since trial runs plateaued there).\textsuperscript{\textdagger} indicates 99\% significance.}
   \label{tab:multilingual-model-size-f1}
 \end{table*}

\section{Effect of Sequence Length on Performance}
\label{sec:effect-of-seq-len}


The results in Table~\ref{tab:multilingual-seq-len} help us understand the robustness of different formats to changes in sequence lengths.
In general, we found that that all the formats were fairly robust to reductions in training sequence length, with previous trends among formats holding, 
In most cases, the performance dropped by about 1-2pp, with the biggest drop observed for the \textbf{Tagged Spans} approach on the Joint setting (about \textbf{-5pp} across metrics).
Thus we can conclude that while setting more appropriate sequence lengths can increase performance, failure to do so does not hurt performance significantly.

\section{Illustrative Hallucination Examples}
\label{sec:hallucination-evals}

Section~\ref{sec:hallucinations} discussed the increased robustness of the \textbf{Sentinel+Tag} model.

To help understand what hallucinations would look like, we provide some examples below using the running example of "Add Kent James to the Disney soundtrack". The \emph{expected Target} (using the \textbf{Tagged Spans } format) is "<O> Add </> <ARTIST> Kent James </> <O> to </> <O> the </> <PLAYLIST> Disney </> <O> soundtrack </>". 

Examples of predictions containing \emph{hallucinations} would be:
\begin{itemize}
    \item "<O> Add </> <ARTIST> Kent Jackson </> <O> to </> <O> the </> <PLAYLIST> Disney </> <O> soundtrack </>" changes the name of the artist from "Kent James" to "Kent Jackson".
    \item "<O> Add </> <ARTIST> Kent James </> <O> to </> <O> the </> <O> soundtrack </>" drops the "Disney" token altogether.
    \item "<O> Add </> <ARTIST> Kent James </> <O> to </> <O> the </> <PLAYLIST> Walt Disney </> <O> soundtrack </>" adds a spurious token "Walt".
\end{itemize}

The above examples can also be trivially modified for the \textbf{Input + Tag} format.
In particular, the first example of the above three hallucinations is also an example of why the metrics for the \textbf{Input + Tag} format (as in Table~\ref{tab:multilingual-results-zs-full} and Table~\ref{tab:multilingual-results-full}) may be a tad generous.
Since the token indices would not have caught the hallucination in the generated token text ("Kent Jackson" instead of "Kent James"), the model's performance drops when looking at the actual text as shown in the metrics in parentheses in Table~\ref{tab:multilingual-hallucination}.

\begin{table*}[h]
\vspace{-0.1in}
\centering
 \resizebox{\textwidth}{!}{
 \begin{tabular}{|c|}
  \hline

\textbf{Input:} Will the temps be freezing tonight
\\
\textbf{Target:} <WEATHER\_ATTRIBUTE> freezing <DATE\_TIME> tonight
\\
\textbf{E TS:} <DATE\_TIME> tonight
\\
\textbf{Sentinel Target = E S+T(S):} <extra\_id\_4> WEATHER\_ATTRIBUTE <extra\_id\_5> DATE\_TIME
 \\ \hline

\textbf{Input:} tonights projected low
\\
\textbf{Target:} <DATE\_TIME> tonights
\\
\textbf{E TS:} <MUSIC\_ALBUM\_TITLE> tonights projected low
\\
\textbf{Sentinel Target = E S+T(S):} <extra\_id\_0> DATE\_TIME
 \\ \hline
 
 \textbf{Input:} what are some ways to cook chicken in the crock pot
\\
\textbf{Target:} <RECIPES\_INCLUDED\_INGREDIENT> chicken <RECIPES\_COOKING\_METHOD> crock pot
\\
\textbf{E TS:} <RECIPES\_DISH> chicken in the crock pot
\\
\textbf{Sentinel Target = E S+T(S):} <extra\_id\_6> RECIPES\_INCLUDED\_INGREDIENT <extra\_id\_9> RECIPES\_COOKING\_METHOD <extra\_id\_10>
 \\ \hline
 
 \textbf{Input:} What is a low carb food
\\
\textbf{Target:} <RECIPES\_QUALIFIER\_NUTRITION> low <RECIPES\_TYPE\_NUTRITION> carb
\\
\textbf{E TS:} <RECIPES\_QUALIFIER\_NUTRITION> low <RECIPES\_TYPE\_NUTRITION> carb
\\
\textbf{Sentinel Target = E S+T(S):} <extra\_id\_3> RECIPES\_QUALIFIER\_NUTRITION <extra\_id\_4> RECIPES\_TYPE\_NUTRITION
 \\ \hline
 
 \textbf{Input:} Read the top story on CNN.com
\\
\textbf{Target:} <NEWS\_REFERENCE> top <NEWS\_TYPE> story <NEWS\_SOURCE> CNN.com
\\
\textbf{E TS:} 
\\
\textbf{Sentinel Target = E S+T(S):} <extra\_id\_2> NEWS\_REFERENCE <extra\_id\_3> NEWS\_TYPE <extra\_id\_5> NEWS\_SOURCE
 \\ \hline
 
 \textbf{Input:} set timer for 28 minutes for cake to be done
\\
\textbf{Target:} <METHOD\_TIMER> timer <DATE\_TIME> for 28 minutes
\\
\textbf{E TS:} <METHOD\_TIMER> timer <DATE\_TIME> for 28 minutes <TIMER\_NAME> cake
\\
\textbf{Sentinel Target = E S+T(S):} <extra\_id\_1> METHOD\_TIMER <extra\_id\_2> DATE\_TIME <extra\_id\_3> <extra\_id\_4>
 \\ \hline
 
 \textbf{Input:} what is the latest new york post headlines
\\
\textbf{Target:} <DATE\_TIME> latest <NEWS\_SOURCE> new york post <NEWS\_TYPE> headlines
\\
\textbf{E TS:} <DATE\_TIME> latest <NEWS\_TOPIC> new york post <NEWS\_TYPE> headlines
\\
\textbf{Sentinel Target = E S+T(S):} <extra\_id\_3> DATE\_TIME <extra\_id\_4> NEWS\_SOURCE <extra\_id\_5> <extra\_id\_6> <extra\_id\_7> NEWS\_TYPE
 \\ \hline

 \end{tabular}
 }

 \caption{Examples of wins for \textbf{Extractive Sentinel+Tag (S)} vs the \textbf{Extractive TaggedSpans} on mTOP.}
 
 \vspace{-0.1in}
   \label{tab:wins-examples}
 \end{table*}

\begin{table*}[h]
\vspace{-0.1in}
\centering
 \resizebox{\textwidth}{!}{
 \begin{tabular}{|c|}
  \hline

\textbf{Input:} are there clear skies tonight
\\
\textbf{Target = E TS:} <WEATHER\_ATTRIBUTE> clear <DATE\_TIME> tonight
\\
\textbf{Sentinel Target:} <extra\_id\_2> WEATHER\_ATTRIBUTE <extra\_id\_4> DATE\_TIME
\\
\textbf{E S+T(S):} <extra\_id\_3> WEATHER\_ATTRIBUTE <extra\_id\_4> DATE\_TIME

 \\ \hline
 
\textbf{Input:} Can you play metal radio from Spotify for me
\\
\textbf{Target = E TS:} <MUSIC\_GENRE> metal <MUSIC\_TYPE> radio <MUSIC\_PROVIDER\_NAME> Spotify
\\
\textbf{Sentinel Target:} <extra\_id\_3> MUSIC\_GENRE <extra\_id\_4> MUSIC\_TYPE <extra\_id\_6> MUSIC\_PROVIDER\_NAME
\\
\textbf{E S+T(S):} <extra\_id\_3> MUSIC\_PLAYLIST\_TITLE <extra\_id\_4> MUSIC\_TYPE <extra\_id\_6> MUSIC\_PROVIDER\_NAME

 \\ \hline
 
 \textbf{Input:} Are there any severe weather advisories for the Pacific Northwest
\\
\textbf{Target = E TS:} <LOCATION> Pacific Northwest
\\
\textbf{Sentinel Target:} <extra\_id\_8> LOCATION <extra\_id\_9>
\\
\textbf{E S+T(S):} <extra\_id\_7> LOCATION <extra\_id\_8> <extra\_id\_9>

 \\ \hline
 
\textbf{Input:} Any science related news
\\
\textbf{Target = E TS:} <NEWS\_CATEGORY> science <NEWS\_TYPE> news
\\
\textbf{Sentinel Target:} <extra\_id\_1> NEWS\_CATEGORY <extra\_id\_3> NEWS\_TYPE
\\
\textbf{E S+T(S):} <extra\_id\_1> NEWS\_CATEGORY <extra\_id\_2> <extra\_id\_3> NEWS\_TYPE

 \\ \hline

\textbf{Input:} i want the current headlines across US
\\
\textbf{Target = E TS:} <DATE\_TIME> the current <NEWS\_TYPE> headlines <NEWS\_TOPIC> US
\\
\textbf{Sentinel Target:} <extra\_id\_2> DATE\_TIME <extra\_id\_3> <extra\_id\_4> NEWS\_TYPE <extra\_id\_6> NEWS\_TOPIC
\\
\textbf{E S+T(S):} <extra\_id\_3> DATE\_TIME <extra\_id\_4> NEWS\_TYPE <extra\_id\_6> NEWS\_TOPIC

 \\ \hline

\textbf{Input:} is there an update on the news story about Bill Clinton
\\
\textbf{Target = E TS:} <NEWS\_TYPE> update <NEWS\_TYPE> news story <NEWS\_TOPIC> Bill Clinton
\\
\textbf{Sentinel Target:} <extra\_id\_3> NEWS\_TYPE <extra\_id\_6> NEWS\_TYPE <extra\_id\_7> <extra\_id\_9> NEWS\_TOPIC <extra\_id\_10>
\\
\textbf{E S+T(S):} <extra\_id\_3> NEWS\_TYPE <extra\_id\_6> NEWS\_TYPE <extra\_id\_7> <extra\_id\_9> NEWS\_TOPIC <extra\_id\_10>

 \\ \hline

 \end{tabular}
 }

 \caption{Examples of losses for \textbf{Extractive Sentinel+Tag (S)} vs the \textbf{Extractive TaggedSpans} on mTOP.}
 
 \vspace{-0.1in}
   \label{tab:losses-examples}
 \end{table*}

\section{Real Hallucination Examples (for Extractive formats)}
\label{sec:hallucination-examples-extractive}

To help us understand the kinds of hallucinations we observe in these models, below are different sets of examples of hallucinations we see in the different formats. While these examples are for the shorter (more succinct) extractive formats they are also indicative of the hallucinations we see for the longer formats that label every input token. Note though that the ordering is not indicative of the prevalence of the different types of hallucinations but largely for the sake of exposition.

\subsection{(Extractive) Tagged Spans}

\textbf{Incorrectly copied or added or modified words / phrases:} This is fairly common even on English datasets.
\\
\\
\textbf{Dataset:} ATIS \\
\textbf{Input:} what is mci \\
\textbf{Label:} <airport\_code> mci \\
\textbf{Prediction:} <airport\_code> mco \\
\\
\textbf{Dataset:} mTOP \\
\textbf{Input:} High and low temps please \\
\textbf{Label:}  \\
\textbf{Prediction:} <DATE\_TIME> High and low <METHOD\_TIMER> time \\
\\
\textbf{Dataset:} mTOD \\
\textbf{Input:} add title ' trash day ' to 8a alarm \\
\textbf{Label:} <datetime> to 8a \\
\textbf{Prediction:} <datetime> today <datetime> to 8a \\
\\
\textbf{Dataset:} SNIPS \\
\textbf{Input:} name a science fiction film from 1961 \\
\textbf{Label:} <GENRE> science fiction <YEAR> 1961 \\
\textbf{Prediction:} <GENRE> science fiction <YEAR> 1960 \\

\textbf{Arbitrarily hallucinated  words / phrases:} Worryingly, this is very common in all languages (including English).
\\
\\
\textbf{Dataset:} ATIS \\
\textbf{Input:} list airports \\
\textbf{Label:}  \\
\textbf{Prediction:} <city\_name> dallas \\
\\
\textbf{Dataset:} mTOD \\
\textbf{Input:} will it pour ? \\
\textbf{Label:} <weather/attribute> pour \\
\textbf{Prediction:} <weather/attribute> para \\
\\
\textbf{Dataset:} mTOD \\
\textbf{Input:} set reoccurring alarms \\
\textbf{Label:}  \\
\textbf{Prediction:} <reminder/noun> reminders \\

\textbf{Translated words / phrases:} An interesting pattern we observe on some non-English datasets is where a word is accidentally translated or transliterated.
\\
\\
\textbf{Dataset:} mTOD \\
\textbf{Input:} apagar todas mi alarmas hoy \\
\textbf{Label:} <datetime> hoy \\
\textbf{Prediction:} <datetime> today \\
\\
\textbf{Dataset:} mTOP \\
\textbf{Input:} Wie hoch ist die Temperatur \\
\textbf{Label:}  \\
\textbf{Prediction:} <RECIPES\_ATTRIBUTE> temperature \\
\\
\textbf{Dataset:} mTOP \\
\textbf{Input:} quel est le pourcentage de chances de pluie pour aujourd'hui \\
\textbf{Label:} <WEATHER\_ATTRIBUTE> pluie <DATE\_TIME> pour aujourd'hui \\
\textbf{Prediction:} <WEATHER\_ATTRIBUTE> rainie <DATE\_TIME> aujourd'hui \\

\subsection{(Extractive) Sentinel+Tag approach}

\textbf{Extra token / tag:} The most frequent (albeit rare) observed is where an extra token is added to the output. 
\\
\\
\textbf{Dataset:} mTOP \\
\textbf{Input:} <extra\_id\_0> Set \\
\textbf{Label:} < \\
\textbf{Prediction:} <extra\_id\_1> METHOD\_TIMER \\
\\
\textbf{Dataset:} mTOP \\
\textbf{Input:} <extra\_id\_0> Invite\\
\textbf{Label:}  \\
\textbf{Prediction:} <extra\_id\_1> CONTACT\_ADDED \\

\textbf{Wrong sentinel number:} Occasionally (and somewhat surprisingly) the model copies the wrong sentinel token index.
\\
\\
\textbf{Dataset:} mTOP \\
\textbf{Input:} <extra\_id\_0> get <extra\_id\_1> events <extra\_id\_2> I'm <extra\_id\_3> going <extra\_id\_4> to \\
\textbf{Label:} <extra\_id\_2> USER\_ATTENDEE\_EVENT \\
\textbf{Prediction:} <extra\_id\_5> USER\_ATTENDEE\_EVENT \\

\textbf{Model expects further tokenization:} Sometimes the models expects the input to have been tokenized more than it was
\\
\\
\textbf{Dataset:} MIT-Movies \\
\textbf{Input:} <extra\_id\_0> who <extra\_id\_1> stars <extra\_id\_2> in <extra\_id\_3> the <extra\_id\_4> movie <extra\_id\_5> titled <extra\_id\_6> happythankyoumoreplease \\
\textbf{Label:} <extra\_id\_6> TITLE \\
\textbf{Prediction:} <extra\_id\_6> TITLE <extra\_id\_7> <extra\_id\_8> \\

\section{Examples of Wins and Losses}
\label{sec:wins-losses-examples}

To help give a sense of the wins and losses we observed using the sentinel approach, we compared the \textbf{Sentinel+Tag} approach vs the current standard \textbf{TaggedSpans} approach on the largest dataset: mTOP.
To make this analysis easier, we analyzed the extractive versions of the models as they are more succinct.
The examples are provided in Tables~\ref{tab:wins-examples} and~\ref{tab:losses-examples}.

In general we find that the sentinel model seems to be generally better at identifying and labeling the spans in the text. However we also find examples where occasionally the sentinel model adds or shortens a multi-word span (e.g. "Pacific Northwest" extended to "the Pacific Northwest", "science" extended to "science related", "the current" shortened to "current").